%% file: main.tex
\RequirePackage{fix-cm}
\documentclass[twocolumn]{svjour3}          % twocolumn
\smartqed  % flush right qed marks, e.g. at end of proof
\usepackage{graphicx}
\usepackage{natbib}
\usepackage{multirow}
\usepackage{amsmath}
\usepackage{amssymb}
\usepackage{bbm}
\usepackage{bm}
\usepackage{booktabs}
\usepackage{array} % for extended column definitions, e.g. >{\em}c
\usepackage{blindtext}
\usepackage{comment}

\usepackage{fancynum}
\usepackage{threeparttable}
\usepackage{subcaption}
\usepackage{soul}
\usepackage{tcolorbox}

\usepackage{inconsolata}
\usepackage{pifont}% http://ctan.org/pkg/pifont
\newcommand{\cmark}{\ding{51}}%
\newcommand{\xmark}{\ding{55}}%

\usepackage{lipsum}
\usepackage{graphicx}
\newcommand{\ie}{\textit{i}.\textit{e}.}
\newcommand{\eg}{\textit{e}.\textit{g}.}
\newcommand{\vs}{\textit{v}\textit{s}.}

\usepackage{color, colortbl}
\definecolor{citecolor}{HTML}{0071bc}
\definecolor{tabhighlight}{HTML}{e5e5e5}
\usepackage[colorlinks,citecolor=citecolor]{hyperref}

\makeatletter
\renewcommand\paragraph{
  \@startsection{paragraph} % name
  {4} % level
  {\z@} % indent
  {.5em \@plus1ex \@minus.2ex} % beforeskip
  {-.5em} % afterskip
  {\normalfont\normalsize\bfseries} % style
}
\makeatother
\begin{document}
\sloppy

\title{VideoQA in the Era of LLMs: An Empirical Study 
\thanks{
This research is supported by the National Research Foundation, Singapore under its NRF Fellowship for AI (NRF-NRFFAI1-2019-0001). Any opinions, findings and conclusions or recommendations expressed in this material are those of the author(s) and do not reflect the views of National Research Foundation, Singapore. 
%about the article that should go on the front page should be
%placed here. General acknowledgments should be placed at the end of the article.
}
}
%\subtitle{Do you have a subtitle?\\ If so, write it here}
%\titlerunning{Short form of title}  % if too long for running head

\author{Junbin~Xiao \and
        Nanxin~Huang \and
        Hangyu~Qin \and
        Dongyang~Li \and
        Yicong~Li \and
        Fengbin~Zhu \and
        Zhulin~Tao \and
        Jianxing~Yu \and
        Liang~Lin \and
        Tat-Seng~Chua \and
        Angela~Yao
}

\institute{
Junbin Xiao, Hangyu Qin, Yicong Li, Fengbin Zhu, Tat-Seng Chua and Angela Yao are with National University of Singapore. 
Corresponding to: ayao@comp.nus.edu.sg.
% E-mail: \{junbin, ayao, chuats\}@comp.nus.edu.sg, hangyu.qin@u.nus.edu liyicong@nus.edu.sg, zhfengbin@gmail.com. 
\\
\indent Nanxin Huang, Dongyang Li and Zhulin Tao are with Communication University of China. 
% E-mail: \{hnancy, lidongyang, taozl\}@cuc.edu.cn. 
Corresponding to: taozl@cuc.edu.cn
\\
\indent Jianxing Yu and Liang Lin are with Sun Yat-Sen University. \\
% E-mail: yujx@mail.sysu.edu.cn, linliang@ieee.org. 
}
\date{Received: date / Accepted: date}
% The correct dates will be entered by the editor

\maketitle

\input{abstract}
\input{intro}
\input{related}

\input{method}

\input{discussion}
\input{conclusion}

\noindent \paragraph{Limitations} 
We note that Video-LLM research is emerging, and there are other powerful models we may have missed testing.
We thus release all our test data: \url{https://github.com/doc-doc/VideoQA-LLMs}.

\noindent \paragraph{Acknowledgements}
We greatly thank OpenAI for offering us with research access API credits.
% for this project.

% BibTeX users please use one of
\bibliographystyle{spbasic}      % basic style, author-year citations
\bibliography{main}   % name your BibTeX data base

\clearpage
\appendix
\input{appendix}
\end{document}

%% file: abstract.tex
\begin{abstract}
Video Large Language Models (Video-LLMs) are flourishing and has advanced many video-language tasks. As a golden testbed, Video Question Answering (VideoQA) plays pivotal role in Video-LLM developing. This work conducts a timely and comprehensive study of Video-LLMs’ behavior in VideoQA, aiming to elucidate their success and failure modes, and provide insights towards more human-like video understanding and question answering. 
Our analyses demonstrate that Video-LLMs excel in VideoQA; they can correlate contextual cues and generate plausible responses to questions about varied video contents. However, models falter in handling video temporality, both in reasoning about temporal content ordering and grounding QA-relevant temporal moments. Moreover, the models behave unintuitively - they are unresponsive to adversarial video perturbations while being sensitive to simple variations of candidate answers and questions. Also, they do not necessarily generalize better. 
The findings demonstrate Video-LLMs' QA capability in standard condition yet highlight their severe deficiency in 
robustness and interpretability, suggesting the urgent need on rationales in Video-LLM developing.

\end{abstract}

%% file: intro.tex
\section{Introduction}\label{sec:intro}
% Large language models (LLMs) are revolutionizing research in various X+Language tasks where X ranges from vision to robotics to sciences, etc. LLMs have ushered in a new era with unprecedented advancements in computer science.
Large language models (LLMs) are revolutionizing research in various X+Language tasks (with X be
vision, robotics, sciences, etc.), ushering in a new era with unprecedented advancements in computer science
\citep{openai2023gpt4,chung2022scaling,touvron2023llama,vicuna2023,dubey2024llama}. Video Question Answering (VideoQA), as a multimodal and 
multifaceted challenge that requires both linguistic and visual reasoning, is at the forefront of this revolution.
It also serves as a litmus test for evaluating the comprehensive 
multimodal understanding and reasoning capabilities of cutting-edge Video-LLMs \footnote{In this paper, we denote Video-LLMs as video-language models that use LLMs with $\geq1$ billion parameters.} \citep{alayrac2022flamingo,dai2023instructblip,maaz2023video,chen2023pali,liu2023visual,zhang2023videollama,lin2023video,li2023llama,gemini2023}. 
\begin{figure*}[!t]
\centering
\includegraphics[width=1.0\linewidth]{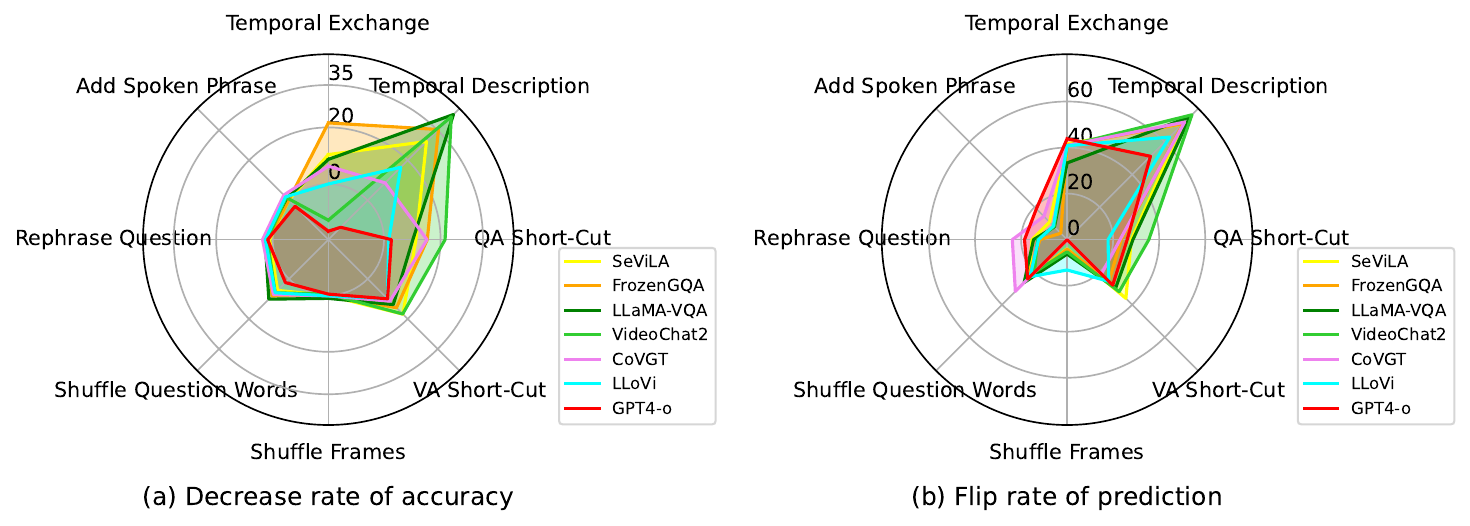}%
\caption{Models' QA behavior under different probes on NExT-QA. The plots show that the models rarely align with human intuition in maintaining their accuracy and specific predictions when facing adversarial challenges.
}
\label{fig:intro}
\vspace{-0.3cm}
\end{figure*}

While VideoQA performance steadily improves with Video-LLMs,
the pursuit of higher QA accuracy alone is insufficient to demonstrate the models' capability, neither in video understanding nor in question-answering. For example, a recent study \citep{xiao2023can} reveals that the models can answer a large portion of questions (correctly) with irrelevant video inputs or even without video at all. 
Also, there is a surge of works \citep{bai2024hallucination} studying the hallucination problem of multimodal LLMs in visual description.
These observations raise the concern about whether these models can provide faithful answers in real-world applications. In this regard, we conduct a comprehensive study to Video-LLM's behavior in VideoQA. We delve into the models' success and failure modes from multiple aspects to provide insights towards more trustworthy and human-like machine intelligence. 

% To this end, 
We begin by collecting a series of representative Video-LLMs \citep{yang2022zero, ko2023large, yu2023self, zhang2023simple, zhang2023videollama, li2023llama, lin2023video,li2023mvbench}.  These models have achieved state-of-the-art (SOTA) on VideoQA in either finetuned or zero-shot settings by harnessing popular LLMs such as DeBERTa-V2 \citep{he2020deberta}, Flan T5 \citep{chung2022scaling}, LLaMA \citep{touvron2023llama}, Vicuna \citep{vicuna2023} and GPT-4 \citep{openai2023gpt4}. The accuracy of these models reaches around 70\%$\sim$80\% on popular benchmarks such as NExT-QA \citep{xiao2021next} and MSVD-QA \citep{xu2017video}, for both multi-choice (MC) QA and open-ended (OE) QA.

With the models, we design a comprehensive suite of adversarial probes to dissect their behavior in either video understanding or question answering. Specifically, the probes are derived by adjusting the original VQA data or settings with targeted tests on: 1) temporal understanding, 2) visual grounding, 3) multimodal VQA reasoning, 4) robustness, and 5) generalization. More details are in Sec.~\ref{sec:method}). The analyses are thus based on a comparison of the model accuracy and specific prediction before and after the adjustment.
For better comparison, we also provide the behavior of the SOTA non-LLM methods 
\citep{fu2023empirical, xiao2023covgt} 
that fully fine-tune small language models
such as BERT \citep{devlin2018bert} and RoBERTa \citep{liu2019roberta}. 

Fig.~\ref{fig:intro} presents an overview of the models' behavior when confronted with some typical probes. 
It shows that the models often violate human intuition in maintaining or changing their accuracy and specific predictions. 
% For example, all models tend to change their answers on invariant probes but maintaned
While no single model wins on all tests, we find that
GPT-4o is fairly among the strongest for answering video questions though it is mainly developed for image understanding. 
% While the behavior often varies among individual models, 
We summarize and outline some of the common findings below, 
more specific analyses are presented in Sec.~\ref{sec:method}: 
% \vspace{-0.3cm}
\begin{enumerate}
    \item \textbf{Temporal Understanding:} Most Video-LLMs answer temporal questions with high accuracy, but they can hardly reason the order of the video content and are even inferior to non-LLM methods in this regard. Notably, GPT-4o shows strong temporal reasoning capabilities.
    
    \item \textbf{Visual Grounding:} Video-LLMs significantly outperform non-LLM methods in answering video questions, but they win marginally in answer grounding. This indicates that their much better performances are largely due to their strength in capturing language priors and spurious vision-text correlations.
    
    \item \textbf{Multimodal VQA Reasoning:} Compared with non-LLM methods, Video-LLMs are better at exploiting the short-cuts in candidate answers (especially video-answer short-cuts) for multi-choice QA, reflecting their deficiency in faithful reasoning from video-question to the correct answers. 

    \item \textbf{Robustness:} Video-LLMs are ``overly robust'' and 
    unresponsive to data perturbation on videos (\eg, shuffling video frames) while being unexpectedly sensitive to language variations (\eg, rephrasing questions), especially for open-ended QA. 
    
    \item \textbf{Generalization:} Fine-tuned Video-LLMs still favour the answers of high frequency as predictions in OEQA. While they generalize better across question types and datasets, this is not guaranteed for each specific model. Additionally, according to our observation on NExT-OOD \citep{zhang2023next}, out-of-distribution (OOD) data for non-LLM methods may not be actual OOD for Video-LLMs.
   
\end{enumerate}

On the one hand, the above findings demonstrate Video-LLM's strengths in performing normal video question-answering. On the other hand, they highlight significant limitations in faithful visual reasoning and conditioning their predictions on the visual content. This underscores the necessity of rationales in developing future Video-LLMs.

%% file: related.tex
\section{VideoQA Background}\label{sec:related_work}

\subsection{Preliminaries} 

VideoQA is the task of answering visual content-related questions for videos. It %includes and 
extends image VQA~\citep{antol2015vqa, goyal2017making} to dynamic scenarios, such as the movement state of objects (\eg, ``\texttt{slow}'' or ``\texttt{fast}''?, ``\texttt{put down}'' or ``\texttt{take away}''?), action repetitions and their transitions over time \citep{jang2017tgif,jang2019video}. VideoQA is studied through multiple-choice (MCQA) and open-ended (OEQA) questions. MCQA provides candidate answers along with each question for models to select the correct one \citep{jang2017tgif,xiao2021next}. 
It is favored in answering inference- and explanatory-type questions (\eg, ``\texttt{why and how}'') that require relatively longer answers. %whose answers are relatively longer. 
OEQA treats each answer as a category and requires the models to predict a correct answer from a predefined answer list \citep{xu2017video, yu2019activitynet}. It is mainly adopted to assess factoid QA where the questions pertain to visual recognition (\eg, ``\texttt{what is}''). We refer to the recent survey \citep{zhong2022video} for a more detailed introduction.

\subsection{VideoQA Technique Evolution}
The VideoQA techniques generally follow the corresponding development of vision and text representation models. The development can be roughly divided into three stages. The first is \textbf{1) CNN + RNN} \citep{jang2017tgif,xu2017video,le2020hierarchical,li2022invariant,xiao2022hqga} where video and text are encoded by 
convolutional neural networks (\eg, ResNet~\citep{he2016deep}) and recurrent neural networks (\eg, LSTM 
~\citep{hochreiter1997long}) respectively. The second stage \textbf{2) CNN/ViT+BERT}~\citep{sun2019videobert,li2020hero,seo2021look,lei2021less,yang2021just,xiao2022vgt,xiao2023covgt,fu2023empirical,li2023discovering,li2023transformer}, features improved visual \citep{dosovitskiy2020image} and language representations \citep{devlin2018bert,liu2019roberta}, where the models also benefit from self-supervised cross-modal pre-training at scale and fine-tuned small language models (\eg, BERT and RoBERTa) on the target datasets.  Finally, the third (and the current) stage \textbf{3) CLIP+LLM}~\citep{yang2022zero,liu2023visual,maaz2023video,li2023blip,yu2023self,ko2023large,zhang2023videollama,dai2023instructblip,zhang2023llamaadapter,lin2023video} features 
pre-trained cross-modal visual encoders (\eg, CLIP-ViT \citep{radford2021learning}) and frozen LLMs (\eg, Flan T5 \citep{chung2022scaling}, LLaMA \citep{touvron2023llama}) with instruction tuning of projection and adaptation modules. 
In the transition of each stage, there are significant performance jumps. Especially in the LLM stage, 
leading models like GPT-4V/o \citep{openai2023gpt4} and Gemini \citep{gemini2023} rival human performance in answering standard visual questions. 

\subsection{LLMs for VideoQA}
Existing LLM-based VideoQA methods can be roughly classified into 3 groups:
\textbf{1) General-purpose MLLMs}, where models are not specialized to VideoQA and are instead capable of handling various vision-language tasks such as image/video QA and captioning. Typical models are InstructBLIP \citep{dai2023instructblip}, Video-ChatGPT \citep{maaz2023video}, Video-LLaMA~\citep{zhang2023videollama}, VideoChat \citep{li2023videochat}, VideoChat2~\citep{li2023mvbench}, Video-LLaVA \citep{lin2023video}, LLaMA-VID~\citep{li2023llama}, and LLaVA-NeXT \citep{zhang2024llavanextvideo}, and commercial models such as GPT-4V/4o \citep{openai2023gpt4}, Gemini \citep{gemini2023}. These models do not finetune on the target datasets but are instruction tuned and prompted to perform different tasks, \eg~VideoQA. 
We refer to the recent survey \citep{tang2023video} for a more detailed introduction.

\textbf{2) Specialized Video-LLMs} exploit frozen LLMs or general-purpose image MLLMs (\eg, BLIP-2 \citep{li2023blip}, LLaVA 
 \citep{liu2023visual}, LLaMA-Adapter \citep{zhang2023llama}) while finetuning adaptor or LoRA \citep{hu2021lora} modules on target VideoQA datasets.
% image MLLMs on target video data, where 
Successful examples include FrozenBiLM \citep{yang2022zero}, SeViLA \citep{yu2023self}, LLaMA-VQA \citep{ko2023large}. Additionally, Video-LLaMA \citep{zhang2023videollama} is a general-purpose Video-LLM, but in this paper we study its finetuned version \citep{xiao2023can} that is similar to LLaMA-Adapter but with Video Q-Former to encapsulate short clips of a video.

\textbf{3) Tool-based Video-LLMs}. These methods are parameter-free and focused on the use or collaboration of different MLLMs (LLaVA, GPT-4V, LaViLA \citep{zhao2023learning}) and LLMs (LLaMA, GPT-3.5 and GPT-4) as tools (\eg, planning, captioning, and question-answering) for VideoQA. Representative examples include Socratic Models \citep{zengsocratic},  ViperGPT \citep{suris2023vipergpt}, MoReVQA \citep{min2024morevqa}, LLoVi \citep{zhang2023simple}, IG-VLM \citep{kim2024image}, TraveLER \citep{shang2024traveler}, VideoAgent \citep{wang2024videoagent, fan2024videoagent} and VideoTree \citep{wang2024videotree}. 
These methods are training-free and perform zero-shot VideoQA. 

\subsection{Empirical VQA Study}
Existing literature analyzed image VQA with non-LLM techniques (\eg, RNNs or BERT) and focus on specific issues such as data imbalance~\citep{goyal2017making,kervadec2021roses}, robustness~\citep{shah2019cycle}, or language bias \citep{niu2021counterfactual}. While~\citep{agrawal2016analyzing} analyze the behavior of VQA models from multiple aspects of generalization to long-tailed data as well as complete image and question understanding, they focus on image QA and the techniques examined are from the CNN+RNN stage. In this paper, we will revisit some of the probes (specified in Sec.~\ref{sec:robust}) and check if the problems get alleviated in the LLM era.

To our best knowledge, no previous work has comprehensively studied Video-LLMs' behavior in VideoQA.
Several works explore specific phenomena with models from the stage of CNN/ViT+BERT. For example, \citep{buch2022revisiting} and \citep{lei2023revealing} have revealed the single-frame bias issue, where a large number of questions can be answered with a single video frame. \citep{bagad2023test} have uncovered that existing VLMs (to be distinguished from MLLMs) that fully finetune BERT lack a sense of time. \citep{zhang2023reducing, zhang2023next} have investigated the problem of imbalanced distribution of candidate answers in MCQA, and show that the models tend to exploit the short-cuts in candidate answers for answer prediction. Recently, \citep{xiao2023can} have studied visually grounded QA and disclose that existing VLMs and MLLMs are poor at substantiating their predictions with visual evidence. 

The above works each focuses on a particular aspect of VideoQA; they study either MCQA or OEQA with non-LLM models. 
Furthermore, most of they only mention the problems to induce their own solutions or datasets, without delving into the extent of the problems.
Given the emerging and revolutionary effect of LLMs for question answering, in this paper, we concentrate on LLM-based techniques and present the first comprehensive analysis for Video-LLMs in VideoQA. We study both MCQA and OEQA either for classification or generation, and additionally compare with non-LLM techniques in face of the same probes.

Additionally, we notice that there is a surge of new benchmarks targeting the analysis of Video-LLMs in a broad aspect of video understanding \citep{li2023vitatecs,li2023mvbench,fu2024video,liu2024tempcompass,li2024videovista} (\eg, diverse video categories such as TV/Magic/Fashion shows and related challenges). In this study, we focus on VideoQA of daily activities for its significance towards human-machine interaction.
While we use common names for the probes (\eg, temporal understanding, multimodal reasoning, $\cdots$), the specific probes and data are entirely different. 

\setlength{\tabcolsep}{1.3pt}
\begin{table*}[t!]
\small
\centering
\caption{Tested models from 4 different groups: Specialized, General-Purpose, Tool-Based and Non-LLMs.}
\label{tab:model}
% \vspace{-0.5em}
\begin{threeparttable}
    \scalebox{0.8}{
    \begin{tabular}{llllll}
        \hline\hline
         & Models & Backbone & Video Encoder & Text Encoder \cr
        \hline
        \multirow{5}*{S-LLM}
        &FrozenBiLM \citep{yang2022zero} &- & CLIP \citep{radford2021learning} & DeBERTa-V2-XL (1B) \citep{he2020deberta}\cr
        &FrozenGQA \citep{xiao2023can} & FrozenBiLM &CLIP \citep{radford2021learning} & DeBERTa-V2-XL (1B) \citep{he2020deberta} \cr
        &SeViLA \citep{yu2023self}  & BLIP-2 \citep{li2023blip} & CLIP \citep{radford2021learning} & Flan-T5-XL (3B) \citep{chung2022scaling} \cr
        &LLaMA-VQA \citep{ko2023large} & LLaMA-Adapter \citep{zhang2023llamaadapter}  & CLIP \citep{radford2021learning} & LLaMA (7B) \citep{touvron2023llama} \cr
        \cline{1-5}
        \multirow{4}*{G-LLM}
        &Video-LLaMA \citep{zhang2023videollama}& BLIP-2 \citep{li2023blip} & EVA-CLIP \citep{sun2023eva} & LLaMA (7B) \citep{touvron2023llama} \cr
        &VideoChat2 \citep{li2023mvbench}& BLIP-2 \citep{li2023blip} & UMT-L \citep{li2023unmasked} & Vicuna (7B) \citep{vicuna2023}\cr
        &Video-LLaVA \citep{lin2023video}& LLaVA \citep{liu2023visual} & LangBind \citep{zhu2023languagebind} & Vicuna (7B) \citep{vicuna2023} \cr
        &LLaMA-VID \citep{li2023llama} &- & CLIP \citep{radford2021learning} & Vicuna (7B, 13B) \citep{vicuna2023} \cr
         &GPT-4o & - & unknown & GPT-4 \cr
         \cline{1-5}
         \multirow{1}*{T-LLM}
        & LLoVi \citep{zhang2023simple} & LLaVA, GPT-4 & LangBind \citep{zhu2023languagebind} & GPT-4 \cr
        
         \hline
         \multirow{3}*{N-LLM}
         &VIOLETv2 \citep{fu2023empirical} & VIOLET \citep{fu2021violet}   & VidSwin \citep{liu2022video} & BERT (110M) \citep{devlin2018bert} \cr
         &CoVGT \citep{xiao2023covgt} & VGT \citep{xiao2022vgt}  & ResNet \citep{he2016deep} & RoBERTa (125M) \citep{liu2019roberta} \cr
         &Temp[CLIP] \citep{xiao2023can} & - & CLIP \citep{radford2021learning} & RoBERTa (125M) \citep{liu2019roberta} \cr
         \hline
    \end{tabular}
    }
    % \vspace{-0.3cm}
\end{threeparttable}
\end{table*}

\setlength{\tabcolsep}{19pt}
\begin{table*}[t!]
\small
\centering
\caption{Involved datasets. MCQA and OEQA denote multi-choice and open-ended QA respectively.}
\label{tab:dset}
% \vspace{-0.5em}
\begin{threeparttable}
    \scalebox{0.8}{
    \begin{tabular}{llcccc}
        \hline\hline
        QA Tasks & Datasets & Video Length & \# Answers & \# Videos & \# QA pairs \cr
        \hline
        \multirow{3}*{MCQA}
        & NExT-QA \citep{xiao2021next} & 42s & 5 & 570 & 5K \cr
        & NExT-GQA \citep{xiao2023can} & 40s & 5 & 990 & 5.5K \cr
        & EgoSchema \citep{mangalam2023egoschema} &180s & 5 & 500 & 0.5K  \cr
        & Perception Test \citep{puatruaucean2023perception} &23s & 3 & 5.9K & 19K \cr
         \hline
         \multirow{1}*{OEQA}
         & MSVD-QA \citep{xu2017video} &  10s & \fnum{1853} & 520 &13K \cr
         & ActivityNet-QA \citep{yu2019activitynet} & 180s & \fnum{1000} & \fnum{1000} & 10K \cr
        \hline
    \end{tabular}
    }
    % \vspace{-0.3cm}
\end{threeparttable}
\end{table*}

%% file: method.tex
\section{Probes and Analyses}\label{sec:method}

\subsection{Overview}
We explore the following probes 
% that are 
to examine Video-LLMs' behavior in VideoQA.
The probes are
related to either video understanding or question answering: 
\begin{enumerate}
    \item \textbf{Temporal Understanding.} VideoQA characterizes temporal dynamics compared with image VQA. Hence, our first study is to check if Video-LLMs can (or to what extent) interpret the chronological order of video content by answering paired adversarial temporal questions, \ie, questions elicited by different time words but are about the same video content, \eg, from ``\texttt{A after B}'' to ``\texttt{B before A}''.
    \item \textbf{Visual Grounding.} 
    To discern possible reasons for model failures in following the video content to answer questions, we further study the problem of visual grounding, which is to check whether or to what extent the Video-LLMs' predictions anchored on the \emph{relevant} video content versus irrelevant contexts.
    \item \textbf{Multimodal VQA Reasoning.} 
    Both temporal understanding and visual grounding are evaluated on MCQA because there is no suitable OEQA benchmarks for such tests. As such, we delve deeper into multimodal reasoning to check how much Video-LLMs can reason faithfully from video and question to the correct answer, without influence from short-cuts of candidate answers.
    \item \textbf{Robustness.} Are Video-LLMs robust (like human) to general data perturbation? We further study if model decisions are invariant and equivariant to more general 
    video and question perturbations that are not specific to a particular short-cut problem.
    \item \textbf{Generalizability.} 
    All the above experiments curate new data for testing, which may have probed model generalization to a certain extent. Yet for a better study, we consider two classic generalization problems of long-tailed and out-of-distribution (OOD). 
    Long-tailed is to predict long-tailed answers within the same dataset. OOD is to answer questions of completely different types or related to different kinds of videos (\eg, short to long videos, 3rd- to 1st-person view videos) across different datasets.
\end{enumerate}

\subsection{Tested Models}
We select prominent Video-LLMs that have reported state-of-the-art performance on VideoQA and summarize them in Tab.~\ref{tab:model}. These models exploit cross-modal pretrained visual encoders (\eg, CLIP \citep{radford2021learning}, EVA-CLIP \citep{sun2023eva} and UMT-L \citep{li2023unmasked}) and powerful LLMs (\eg, DeBERTa-V2 \citep{he2020deberta}, Flan T5 \citep{chung2022scaling}, LLaMA \citep{touvron2023llama}, Vicuna \citep{vicuna2023}, GPT-4 \citep{openai2023gpt4}) to perform VideoQA. 

Models in the 1st block of Tab.~\ref{tab:model} are specially finetuned on target datasets to achieve VideoQA. Models in the 2nd block are general-purpose Video-LLMs and most of they do a zero-shot evaluation on VideoQA.
As exception, VideoChat2 \citep{li2023videochat} includes the training data of NExT-QA \citep{xiao2021next} (one of our major testbed) for instruction tuning, and thus should be considered a normal evaluation versus zero-shot. Also, for Video-LLaMA, we adopt the implementation in \citep{xiao2023can}, which is fine-tuned on NExT-QA by adding adpation parameters as in LLaMA-Adapter \citep{zhang2023llamaadapter}.
Additionally, we include the recent GPT-4o \citep{openai2023gpt4} as a representative of commercial models for off-the-shelf use. In the 3rd block, we include LLoVi \citep{zhang2023simple} which is a tool-based Video-LLM established on LLaVA \citep{liu2023visual}, LLaViLA \citep{zhao2023learning} and GPT-4. 
% Note that most of the other tool-based methods are not open-sourced.
As non-LLM baselines (the bottom block), we consider VIOLETv2, CoVGT and Temp[CLIP] %are enclosed 
as representative methods for their good performance achieved by fully finetuning small language models, such as BERT and RoBERTa. 

All models except for Video-LLaMA, we use their official checkpoints (or API for GPT-4o). It is noteworthy that different model often tackles a different type of QA, either MCQA or OEQA. For GPT-4o, we uniformly sample video frames and feed them along with the questions into the API and prompt it for QA. Related details are presented in the Appendix \ref{sec:appgpt4o}.
\vspace{-0.3cm}

\subsection{Datasets \& Evaluation Measures}
Our major experiments are conducted on NExT-QA \citep{xiao2021next} for multi-choice QA, and MSVD-QA \citep{xu2017video} for open-ended QA. 
We select the datasets for their high quality and popularity with completed models of both non-LLMs and LLMs. 
Moreover, we add other datasets for specific probes as necessary. Concretely, we test on NExT-GQA \citep{xiao2023can} for visually grounded QA (Sec.~\ref{sec:ground}), NExT-OOD \citep{zhang2023next} for multi-choice biases in multimodal VQA reasoning (Sec.~\ref{sec:mmreason}) as well as generalization (Sec.~\ref{sec:gener}), ActivityNet-QA \citep{yu2019activitynet} for robustness in answering open-ended temporal questions with shuffled frames (Sec.~\ref{sec:robust}). 
To probe generalization across different types of videos, we add EgoSchema \citep{mangalam2023egoschema} and Perception Test \citep{puatruaucean2023perception} datasets (Sec.~\ref{sec:gener}).  The former features long-form video understanding from an ego-centric viewpoint, while the latter 
emphasizes visual perception and reasoning of fine-grained human-object interactions.
\begin{figure*}[t!]
    \centering
  \subfloat[Flip rates. \label{fig:temp-flip-res}]{%
       \includegraphics[width=0.49\linewidth]{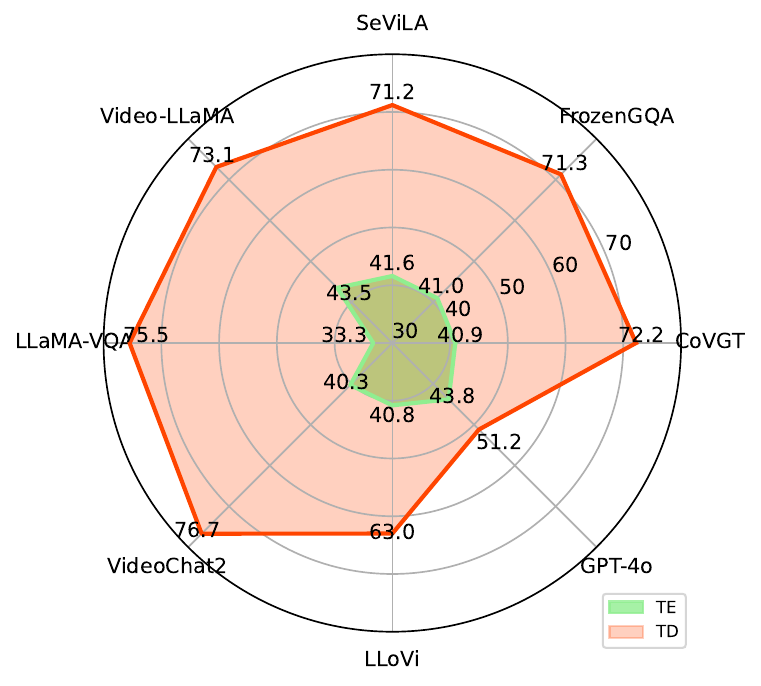}}
    \hfill
  \subfloat[Accuracy. \label{fig:temp-acc-res}]{%
        \includegraphics[width=0.49\linewidth]{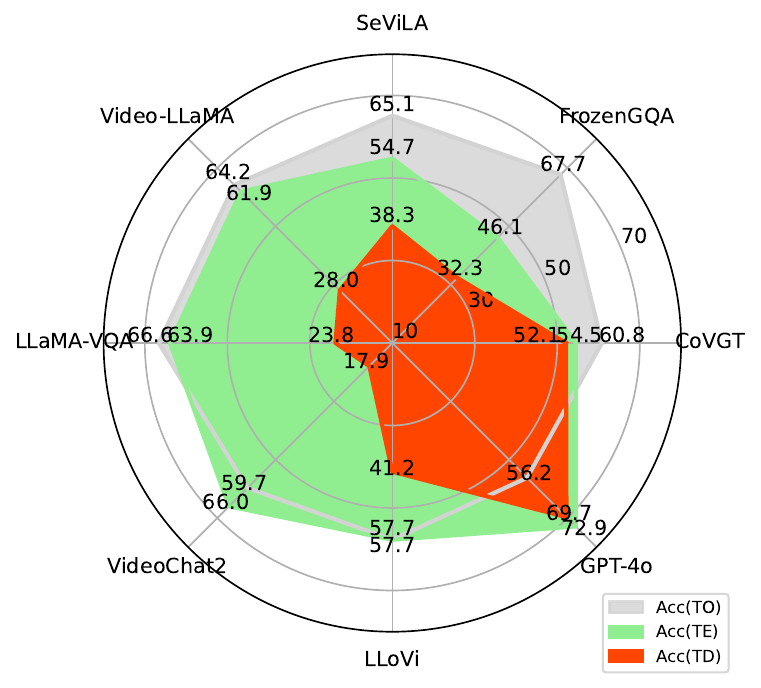}}
  \caption{Results of temporal probing. Most models frequently change their predictions and show decreased accuracy in answering temporally homologous questions w.r.t the chronological order of video contents.}
    \label{fig:temp-res}
    \vspace{-0.4cm}
\end{figure*}
\begin{figure}[t!]
\centering
\includegraphics[width=\linewidth]{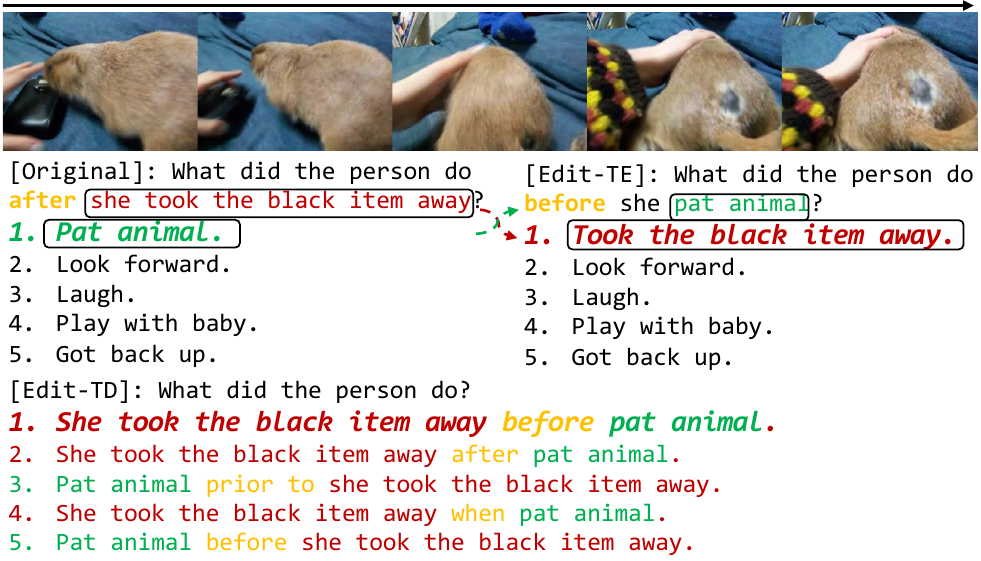}%
\caption{Examples for probing temporal understanding. Correct answers are highlighted in \textbf{\emph{bold and italic}}.}
\label{fig:temp}
\vspace{-0.4cm}
\end{figure}
Other details are listed in Tab.~\ref{tab:dset}. NExT-OOD shares the same statistics as NExT-QA so we do not list it in the table. 
Finally, for evaluation, we report both accuracies and flip rates of specific predictions before and after the probes, where the flip rate denotes the proportion of changed predictions.

\subsection{Temporal Understanding} \label{sec:temp}
VideoQA distinguishes from image VQA for its factoring a rich set of temporal questions signaled by time words ``\texttt{before/after/when}''. Video-LLMs demonstrate increasingly high accuracy in answering temporal questions \citep{xiao2021next}, so we try to verify if these models really interpret content order in videos.
To this end, we derive two new sets of questions from the original temporal questions in NExT-QA.  

Considering the example in Fig.~\ref{fig:temp}. 
The original question asks the person's next action \emph{after} ``\texttt{taking the black item away}'' and should be answered with ``\texttt{pat the animal}''. Our first edit, called \textbf{Temporal Exchange (TE)}, %is to 
exchanges the actions in the question and the %one in the 
correct answer, by asking for the action \emph{before} ``\texttt{the person pats the animal}'' (right of Fig.~\ref{fig:temp}). The models
should be able to answer ``\texttt{take the black item away}'' for the derived question if it correctly answered the original question. 
A second edit called \textbf{Temporal Description (TD)}, focuses on the time order and isolates the effect of language correlation by asking in a coarse manner of ``\texttt{what did xx do or what is happening/doing}". The bottom of Fig.~\ref{fig:temp} shows a TD example, where the question provides little to no contextual information compared to the original 
and TE questions. Moreover, all answers for the question describe the same visual content with different orders though only one is correct.  
Correctly answering TD questions requires stronger action order reasoning compared to the original and TE questions. 
\begin{figure*}[!t]
\centering
\includegraphics[width=\linewidth]{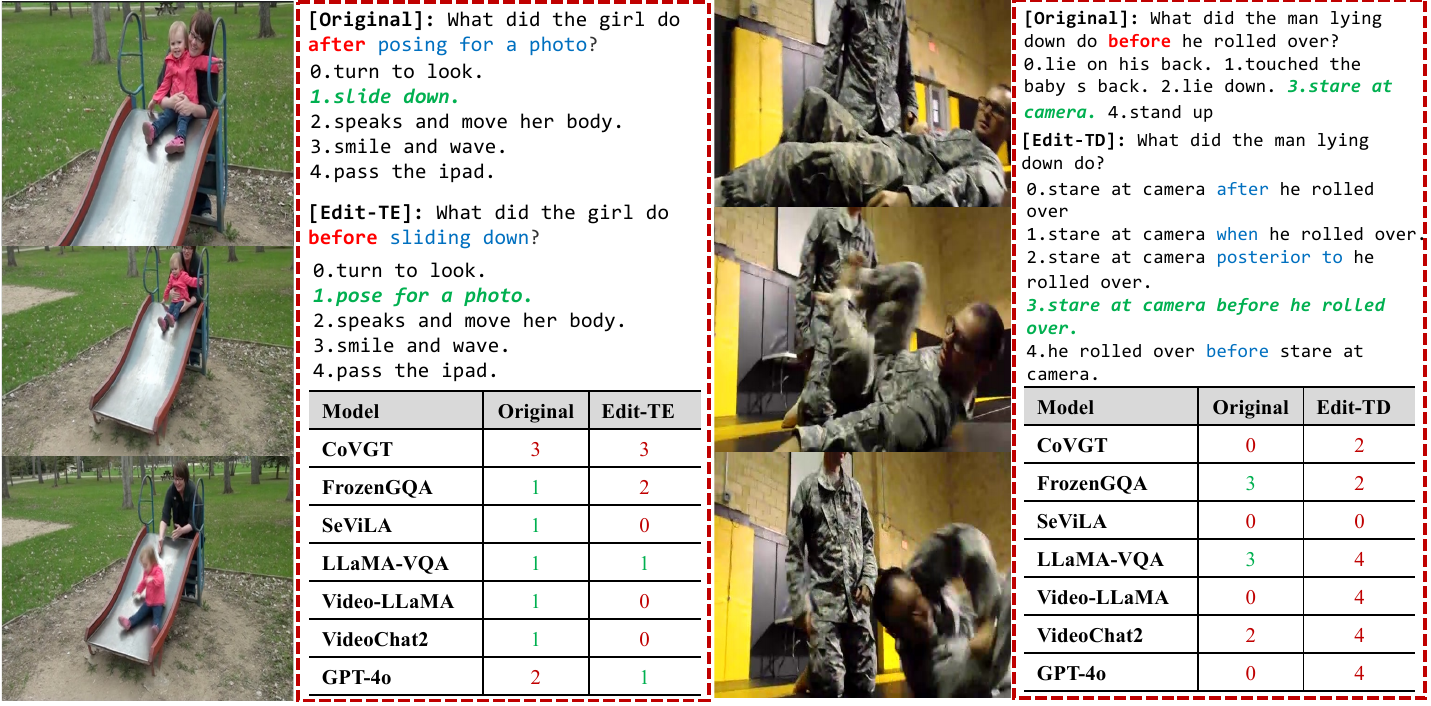}%
\caption{Visualization of temporal probing results. Examples show that most models tend to change their predictions and make mistakes. Correct answers or predictions are in green.}
\label{fig:vis-temp-res}
\vspace{-0.4cm}
\end{figure*}

To obtain the testing examples, we collect about $1K$ temporal questions (TO) originated from NExT-QA and derive the TE and TD examples by prompting GPT-4 and parsing the syntactic structure of the questions respectively. Regarding prompting for TE examples, we split the questions into two groups of ``before/after" and ``when", and design different prompts for them to achieve better results. Regarding parsing for TD questions, we mainly detect the time words ``before/after/when", keep the question parts (\eg, ``\texttt{what did the person do?}") as new questions, and recombine the contextual parts (\eg, ``\texttt{she took the black item away}'') with the correct answers (\eg, ``\texttt{pat animal}") while varying the time words to form different options (see Fig.~\ref{fig:temp}). More details are presented in Appendix \ref{sec:apptemp}. For both the generated TE and TD examples, we further manually check and refine as necessary to ensure the quality of the derived QAs.

The results in Fig.~\ref{fig:temp-res} demonstrate that most models are brittle towards such temporal tests. Specifically, 
the high flip rates in Fig.~\ref{fig:temp-flip-res} show that both LLM and non-LLM methods tend to change their answers for the edited questions. 
Moreover, the accuracy in Fig.~\ref{fig:temp-acc-res} also show that most models' performances shrink.
The effects are specially prominent for the TD test, where most models flip more than 70\% of their predictions and decline their accuracy by more than 30\%, even though the involved video contents remain the same (also refer to prediction examples in Fig.~\ref{fig:vis-temp-res}). 

Among the models, GPT-4o significantly outperforms others in the TD test even though it does not encode any temporal information (we specify in the prompts that the images are ordered video frames). By jointly considering the accuracy in Fig.~\ref{fig:temp-acc-res}, we speculate that the models trained on the target datasets (all models except for GPT-4o and LLoVi \citep{zhang2023simple}) likely overfit to dataset statistics rather than reason about time faithfully. 
Additionally, LLaMA-VQA \citep{ko2023large} shows the smallest flip rate in TE test. We attribute such strength of LLaMA-VQA to its auxiliary training objective by inversely predicting the questions given the answers, which implicitly suits the TE probes by enhancing the relation between the answer and the contextual term in the question. 
Moreover, Fig.~\ref{fig:temp-acc-res} shows that models using advanced LLMs such as LLaMA \citep{li2023llama} and Vicuna \citep{vicuna2023} perform better on the TE test but worse on the TD test than non-LLM method CoVGT. This indicates that LLM-based methods are better at exploiting language priors or commonsense in the TE questions for predicting answers, even though they seem weaker than non-LLM methods in understanding video content order. %the order of the video contents. 
Again, we find that GPT-4o achieves the highest accuracy in both TE and TD tests, showcasing strong zero-shot temporal QA capability.

The above observations, together with the findings in some related studies \citep{bagad2023test, liu2024tempcompass}, suggest that the end-to-end learning approach (even with LLMs) may not be a good choice to cope with temporal relations.  
It seems more favourable to explicitly reason across the sequence of video frames in a way like GPT-4o, and alternatively, to convert the video frames into time-aware descriptions and then feed them into LLMs for question answering, such as in MoReVQA \citep{min2024morevqa} and LLoVi \citep{zhang2023simple}. Yet how to effectively and efficiently coordinate visual captioning and question-answering is still open challenge that deserves further efforts.

\begin{figure*}[t!]
    \centering
  \subfloat[\label{fig:ground_nbp}]{%
       \includegraphics[width=0.4\linewidth]{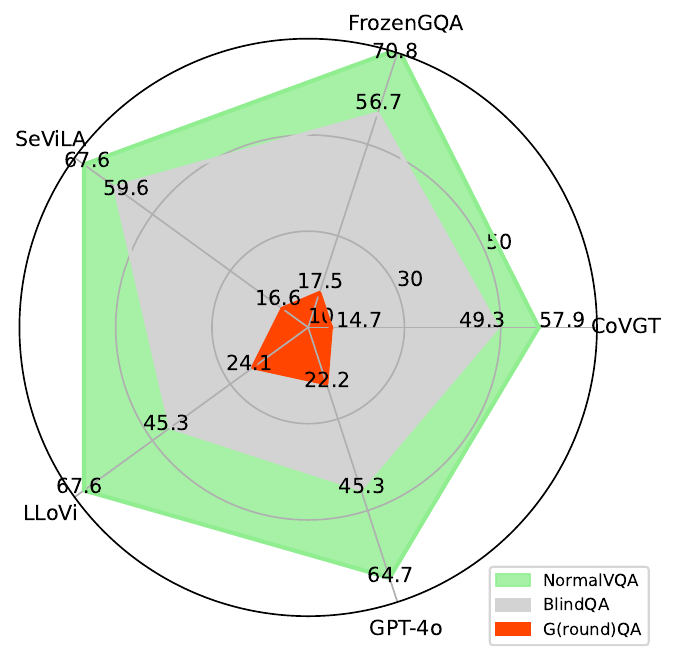}}
    \hfill
  \subfloat[\label{fig:ground_nnp}]{%
        \includegraphics[width=0.4\linewidth]{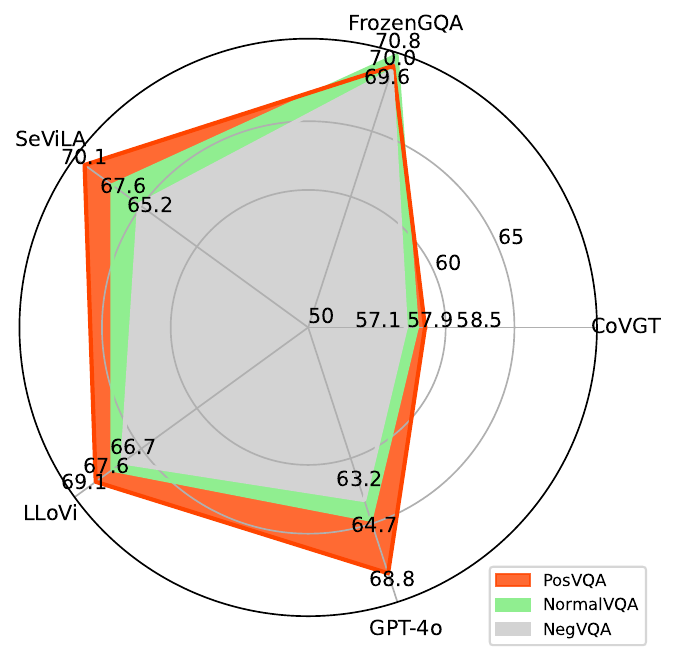}}
  \caption{Results of grounding probing. All models hardly rely on the \emph{relevant video segments} to perform QA though \emph{videos are important} for better performances. Note that a lower BlindQA (or NegVQA) result with also a larger gap with NormalVQA indicates better grounding behavior. }
    \label{fig:ground-res}
    \vspace{-0.4cm}
\end{figure*}
\begin{figure}[!t]
\centering
\includegraphics[width=\linewidth]{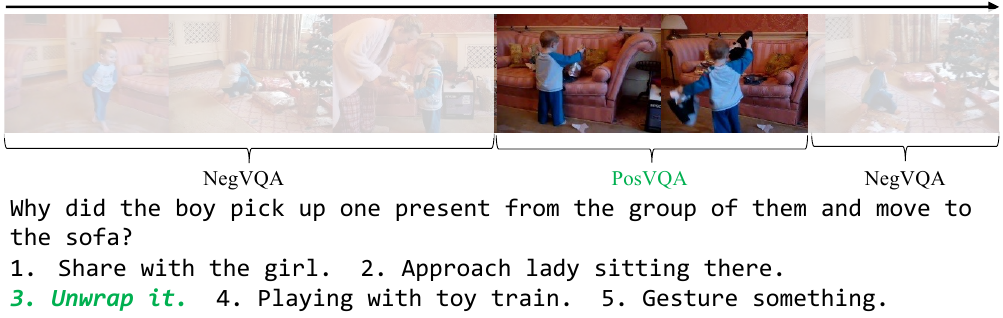}
\caption{Illustration of PosVQA and NegVQA.}
\label{fig:ground}
\vspace{-0.4cm}
\end{figure}
\subsection{Visual Grounding} \label{sec:ground}
With the observation that model answers flip a lot even with paired questions about the same video content, a natural question arises -- to what extent are the models' predictions grounded on the relevant video content? To study this, we conduct experiments on visual grounding. Specifically, we experiment on the recent NExT-GQA  dataset~\citep{xiao2023can}. This dataset extends NExT-QA with temporal location labels of question-answer pairs. 
With NExT-GQA, we compare model performance with different visual inputs as follows:
\textbf{NormalVQA}  
follows the common practice of performing VideoQA by uniformly sampling frames over the whole video. 
% for performing VideoQA. 
\textbf{BlindQA} removes the video inputs to train and test the models with only the QA pairs to check reliance on pure language priors. \textbf{PosVQA} and \textbf{NegVQA} provides only QA-relevant (positive) or only irrelevant segments (negative) as inputs to the model (see Fig.~\ref{fig:ground} as an illustration). 
Finally, we evaluate \textbf{Grounded QA} (GQA) accuracy, \ie, jointly answer the questions and also localize the QA-relevant video moments. By comparing QA performance with and without requirements on visual grounding, we study how much the models' predictions are anchored on the relevant video content. Note that similar experiments are also conducted in NExT-GQA \citep{xiao2023can}, but their analyses are focused on the connection between visual grounding and VQA, with also their major findings established on non-LLM techniques. In this paper, we inherent such settings but target a thorough analysis of Video-LLMs' behavior in visual grounding and a comparison with non-LLM techniques.

The results in Fig.~\ref{fig:ground-res} deliver the following major observations: \textbf{1)} Fig.~\ref{fig:ground_nbp} shows that both LLM and non-LLM methods are poor at substantiating their answers with relevant video content. For example, from NormalVQA to Grounded VQA, \ie~when the correct predictions are required to be visually grounded, the accuracies decrease by more than 40\% for all models.  

\textbf{2)} Compared with the non-LLM method (CoVGT), specialized Video-LLMs (\eg, FrozenGQA and SeViLA) benefit more from short-cut learning. This is because they remarkably surpass the non-LLM method on normal VQA accuracy (by $\sim$10\%) but are only marginally better on GQA (\eg, by $\sim$2\%) (Fig.~\ref{fig:ground_nbp}). Also their LLMs alone (\eg, DeBERTa-V2-XL and Flan-T5-XL) already wins over CoVGT's language model (RoBERTa) by more than 7\% according to the corresponding BlindQA accuracies. A further comparison of NormalVQA \vs~ BlindQA (Fig.~\ref{fig:ground_nbp}), NormalVQA \vs~ PosVQA \vs~ NegVQA (Fig.~\ref{fig:ground_nnp}) indicates that the short-cuts are primarily from language priors and spurious vision-text correlations. For example, the BlindQA baselines without any visual inputs retain more than 70\% of the corresponding NormalVQA' performance. Moreover, visual signals that come from answer-irrelevant video segments yield surprisingly high performance that is competitive to model variants with answer-relevant (ground-truth) visual segments (NegVQA \vs~PosVQA). 

\textbf{3)} Comparing among models, Fig.~\ref{fig:ground_nbp} manifests that the predictions of LLoVi (using GPT-4) and GPT-4o are more visually grounded. Because they achieve the highest GQA accuracy despite being zero-shot methods. Nonetheless, Fig.~\ref{fig:ground_nnp} shows that SeViLA and GPT-4o' predictions are more responsive to different video inputs. Because they have the largest performance gaps between QA with the relevant (positive) and irrelevant (negative) temporal segments.
By comparing the BlindQA performance of the finetuned models (CoVGT, FrozenGQA and SeViLA) and zero-shot models (LLoVi and GPT-4o) in Fig.~\ref{fig:ground_nbp}, we can conclude that finetuned models are much better at exploiting statistic language biases for QA. For example, a finetuned RoBERTa (CoVGT) performs even better than zero-shot GPT-4, 49.3\% \vs~ 45.3\%. 

Additionally, in our implementation, when GPT-4o is prompted to also output the indexes of the video frames that it uses to predict the answers versus just answering the questions, its QA accuracy increases remarkably by more than 15\%, from 49.0\% to 64.7\% for NormalVQA. This suggests GPT-4o's great potential in reasoning like human and providing visually grounded answers if appropriately prompted. We believe this is also an interesting finding and raise it (how to use/prompt MLLMs like GPT-4o) as an open challenge for future exploration.

\subsection{Multimodal VQA Reasoning} \label{sec:mmreason}
\begin{figure}[!t]
\centering
\includegraphics[width=\linewidth]{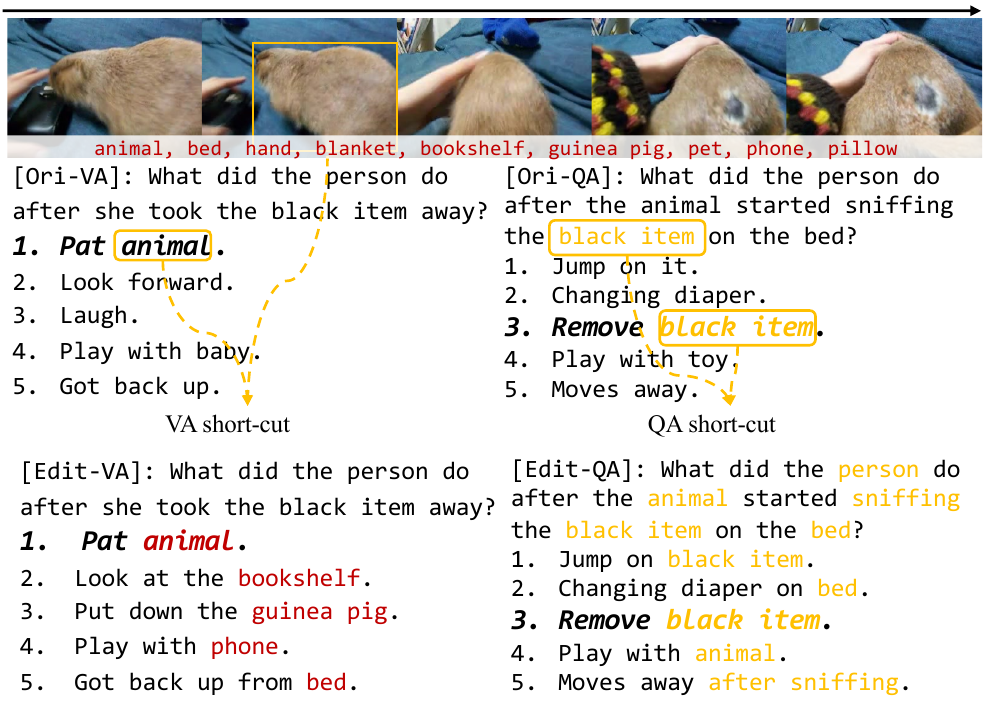}
\caption{Illustration of mitigating the VA (left) and QA (right) short-cuts by incorporating into the negative answers with detected visual concepts and question key words respectively.}
\label{fig:mmreason}
\vspace{-0.5cm}
\end{figure}
The analyses of both temporal understanding and visual grounding are based on MCQA because current OEQA benchmarks do not support such tests, \eg, no sufficient temporal questions or no temporal location labels. Thus, we want to study if (or to what extent) the models' failure of reasoning is because of the short-cuts in candidate answers. We call this probe as multimodal reasoning because such short-cuts dispense with the reasoning between videos and questions, leading to a direct jump to the answers. 

Accordingly, we study two kinds of short-cuts related to candidate answers.
A \textbf{video-answer (VA) short-cut} exists when only one candidate answer contains the visual objects presented in the video, while most other answers do not involve video-related objects. As such, models may ignore the question and choose the video-relevant answer as final prediction. 
A similar \textbf{question-answer (QA) short-cut} exists when only one candidate answer contains question-relevant keywords. This shortcut allows models ignore the video and also bypass interpreting the question as well; they may simply associate common keywords in question and answer for prediction. Fig.~\ref{fig:mmreason} illustrates such VA and QA short-cuts. Note that the short-cuts may not necessarily bring correct answers but can result in false positives, if for example the video-object related answers are not the correct ones. This is possible because the correct answers often contain the actions (or verbs) presented in the videos though without the objects.

\begin{figure}[!t]
\centering
\includegraphics[width=\linewidth]{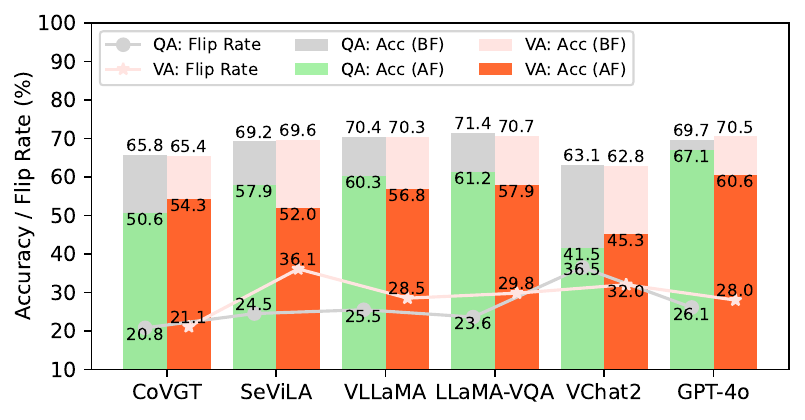}
% \vspace{-0.8cm}
\caption{Results of multimodal VQA reasoning. Acc(BF) and (AF) denote accuracy before and after the edit. VLLaMA and VChat2 are abbreviated from Video-LLaMA and VideoChat2. The flip rates and decrease rate of accuracy demonstrate that the models are susceptible towards such tests.}
\label{fig:mmreason-res}
\vspace{-0.4cm}
\end{figure}

\begin{figure*}[!t]
\centering
\includegraphics[width=\linewidth]{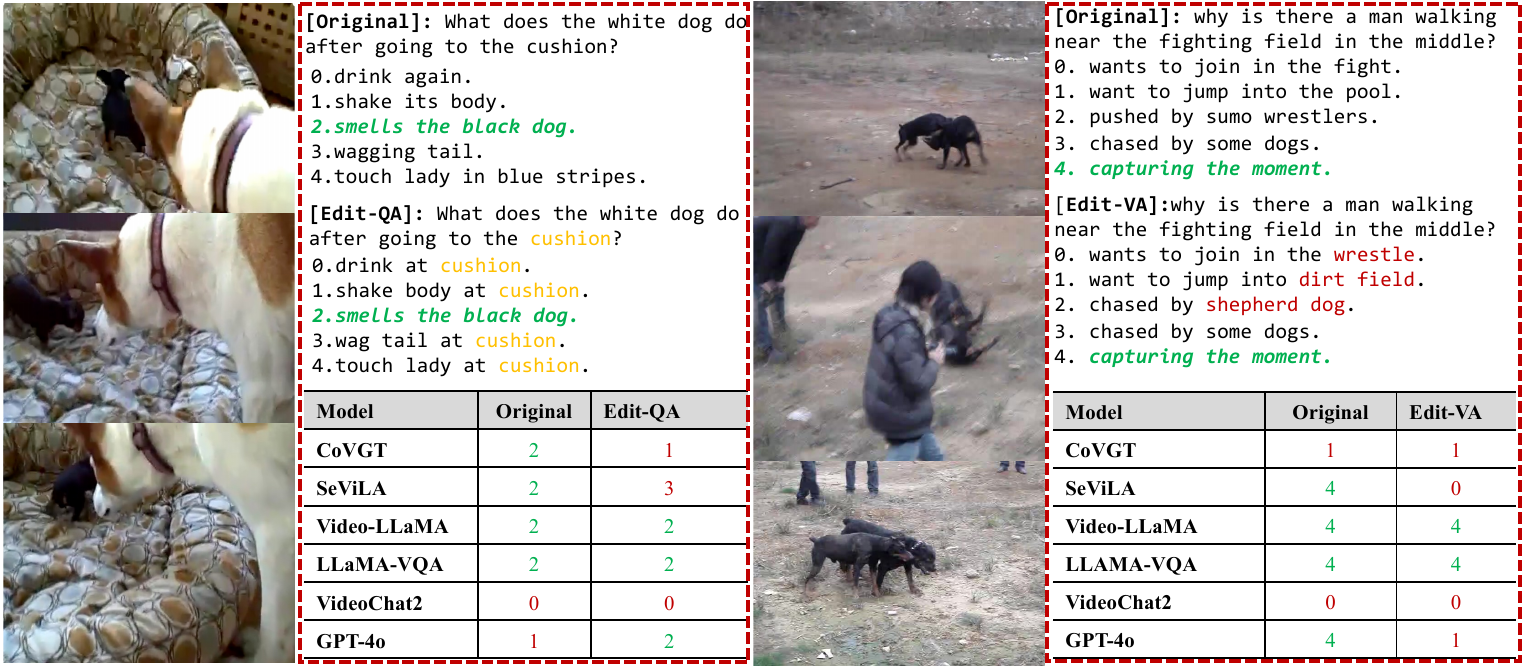}%
\caption{Visualization of predictions before and after removing the QA (left) and VA (right) short-cuts. Examples show that the models are prone to get distracted by the new negative answers.}
\label{fig:vis-mmvqa-res}
\vspace{-0.4cm}
\end{figure*}

To study the problem, we randomly sample 30\% ($\sim$1.2K to save API budgets) of the validation data from NExT-QA. We then check for the samples that have the aforementioned VA and QA short-cuts.
For the target samples, we curate new answers for each question on the basis of the original candidate answers to avoid the possibility of having such shortcuts, \ie, do little change to make most of the candidate answers contain the corresponding video objects or question keywords as shown in the bottom of Fig.~\ref{fig:mmreason}. 
In our implementation, we prompt GPT-4 to achieve the goal, followed by human checking and correction as necessary.

Specifically, to eliminate \textbf{VA short-cut}, we first obtain from the related video a collection of visual object labels. We begin by detecting object labels from the sampled video frames by using image-tagging model RAM \citep{zhang2023recognize} and then filter for redundancy before merging the labels together. With the object labels (Examples in the top of Fig.~\ref{fig:mmreason}) and the original QAs, we then prompt GPT-4 \citep{openai2023gpt4} to incorporate into each candidate answer the appropriate visual label to make it relate to the video (Examples in the bottom left of Fig.~\ref{fig:mmreason}). 

For \textbf{QA short-cut}, we follow a similar practice to prompt GPT-4 to reword each answer option to contain keywords in the question. Note that the identification of samples with short-cuts are completed at the same time by asking GPT-4 to first check before modifying the answers (details in Appendix \ref{sec:appmm}) .
We removed samples whose candidate answers are the same as the original ones as these samples are identified by the GPT-4 as qualified samples without VA and QA short-cuts. Eventually, we obtain two different sets of samples with both size of $1K$ for probing of VA and QA short-cuts respectively. The high short-cut rates ($1K$ out of $1.2K$) is because our short-cuts are defined on nouns while NExT-QA features verbs in its answers.

The results in Fig.~\ref{fig:mmreason-res} show that almost all models, regardless of the language models used, suffer substantial declines in accuracy (10\% $\sim$ 20\%) when answering questions with edited distractor answers (see prediction examples in Fig.~\ref{fig:vis-mmvqa-res}). The flip rates are also high with 21\% $\sim$ 37\% even though the triplets of $<$video, question, correct answer$>$ remain the same. This posits that the models largely rely on the short-cuts in candidate answers and are limited in faithful multimodal reasoning from video and question to correct answers. Again, GPT-4o's zero-shot performance surpasses others on the edited data.

Compared with non-LLM method (CoVGT), all Video-LLMs, especially VideoChat2, show higher flip rates when confronted with the edited options. This suggests that Video-LLMs are better at exploiting the short-cuts of candidate answers for question answering.
By comparing the performances on QA and VA tests, we find that most Video-LLMs experience greater performance loss on edited VA data compared to QA data, especially for GPT-4o whose performance degenerates marginally on QA questions (2.6\%) but more on VA type questions (9.9\%). Such a discrepancy suggests that the models are primarily exploiting spurious vision-text correspondence for answer prediction.
\setlength{\tabcolsep}{4pt}
\begin{table}[t]
	\center
	\small
	\caption{Performance on NExT-OOD Test set (N=1).} 
        % \vspace{-0.3cm}
	\label{tab:mmreason-next-ood-res}
	\scalebox{0.8}{
		\begin{tabular}{l|c|c|ccc}
			\hline\hline
			\multirow{2}{*}{Methods} & \multirow{2}{*}{LLM} &\multicolumn{1}{c|}{NExT-QA} & \multicolumn{3}{c}{NExT-OOD} \\ \cline{3-6}
			&  &  All & VA  &  QA &  VQA \\ \hline
			Prev. SoTA \citep{zhang2023next} & \xmark & 43.7 & 39.6 & 36.6 & 35.8\\
                \hline
 			FrozenGQA \citep{xiao2023can} & \cmark & 73.1 & 71.3 & 71.6 & 72.3 \\  
                SeViLA \citep{yu2023self} & \cmark &  71.8 & 70.1 & 71.4 & 70.0 \\  
            \hline
		\end{tabular}
	}
	\vspace{-0.4cm}
\end{table}

\begin{figure}[t!]
\centering
\includegraphics[width=\linewidth]{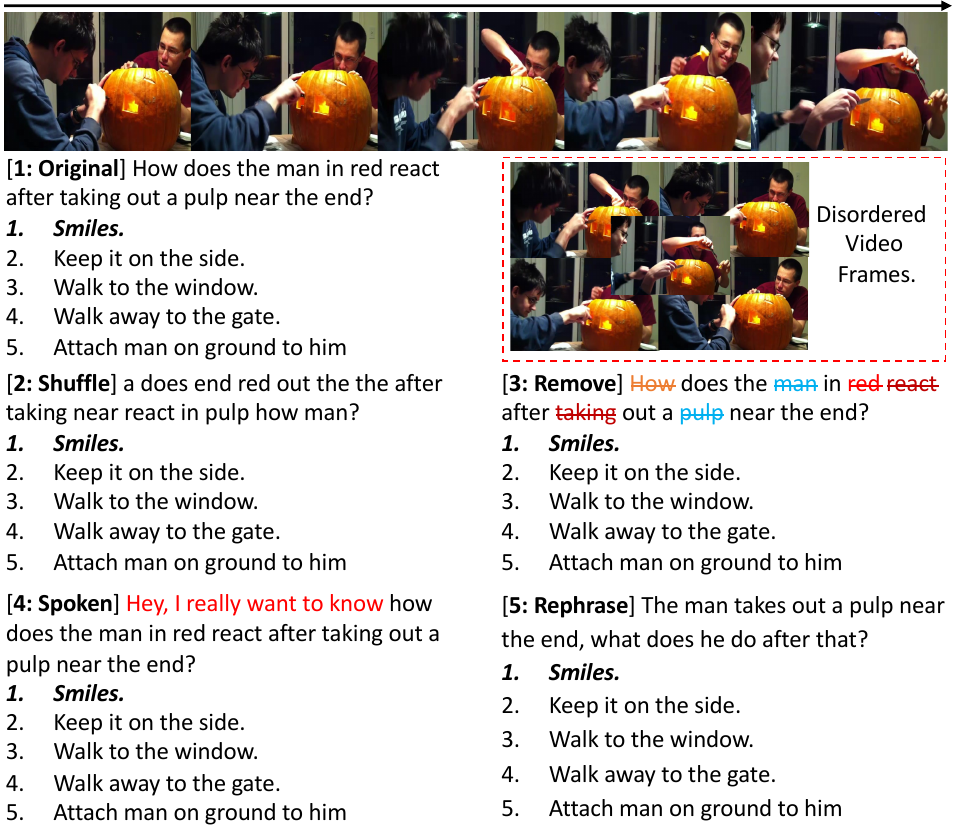}
\caption{Examples of video and question perturbations. }
\label{fig:robust}
\vspace{-0.4cm}
\end{figure}

Additionally, we extend our experiments to the recent NExT-OOD dataset~\citep{zhang2023next}.  
NExT-OOD is also designed to reduce candidate answer biases in NExT-QA. It aims for a balanced distribution of objects and actions in the correct and distractor answers, whereas our VA and QA short-cuts pertains to the video- and question- correlated candidate answers. The results in 
Tab.~\ref{tab:mmreason-next-ood-res} show that Video-LLMs perform surprisingly well on NExT-OOD, surpassing the previous non-LLM SoTA by significant margins.
Moreover, the accuracies of Video-LLMs only slightly decrease from NExT-QA to NExT-OOD. We speculate that NExT-OOD unintentionally introduces new and simple patterns that LLMs can %easily 
exploit, such as mismatches between questions and answers as well as other language issues. This additionally reveals that OOD-data for conventional VideoQA techniques may not be effective OOD for LLM-based methods.

\subsection{Robustness} \label{sec:robust}
The brittle behavior of models in the above studies prompts us to conduct a more general test %about the models' 
on model robustness towards common perturbations on the videos and questions. To this end, we consider two kinds of perturbation: \textbf{data diversifying} and \textbf{data poisoning}, and experiment on both MCQA and OEQA. 

% How resilient are Video-LLMs to video and question perturbations? \hl{Are they able to maintain invariant and equivariant predictions that align with human intuitions?}  
For {\bf data diversifying}, we modify the original questions by rephrasing them or prepending spoken phrases (see examples in the bottom of Fig.~\ref{fig:robust}), both are achieved by prompting GPT-4 with human correction as necessary.
% curate new questions by rephrasing questions and adding spoken phrases (via prompting GPT-4~\citep{openai2023gpt4}) to the original questions , while ensuring the meanings of the questions remain the same. 
Robust models should be invariant and not flip their predictions for this perturbation, since the meanings of questions keep unchanged. 
For {\bf data poisoning}, we attack the input data by shuffling video frames, shuffling question words, and removing keywords from the questions (\eg, question words `6W1H'\footnote{Who, What, When, Where, Why, Which, How}, nouns, and verbs). Examples are presented in the first three cases of Fig.~\ref{fig:robust}. Intuitively, models should suffer significant performance loss with poisoned data.
For data shuffling, we repeat the experiment 3 times with different random seeds and find there is negligible variance in performance ($\pm$ 0.1\%). Therefore, we report only the median results for brevity. % as references. 
Additionally, for data poisoning, we revisit the issue of 
``\emph{answering incomplete questions}'' explored by \citep{agrawal2016analyzing}, analogous to human behavior of answering by listening to only a part of the question. We realize this by progressively truncating the question, starting from the first word and increasing the omitted portion.
\begin{figure*}[t!]
    \centering
  \subfloat[Add spoken words. \label{fig:robust-iv-sp}]{%
       \includegraphics[width=0.49\linewidth]{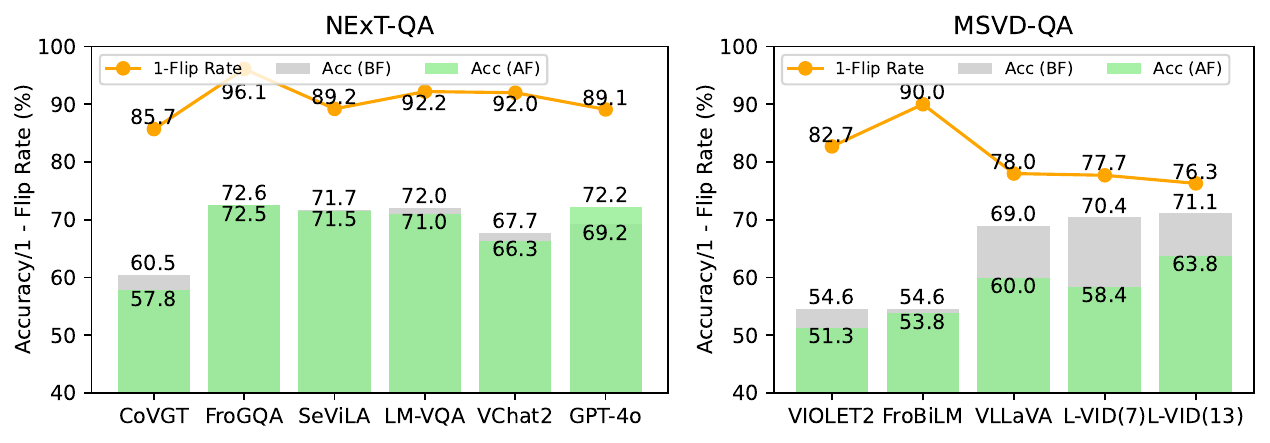}}
    \hfill
  \subfloat[Rephrase questions. \label{fig:robust-iv-rq}]{%
        \includegraphics[width=0.49\linewidth]{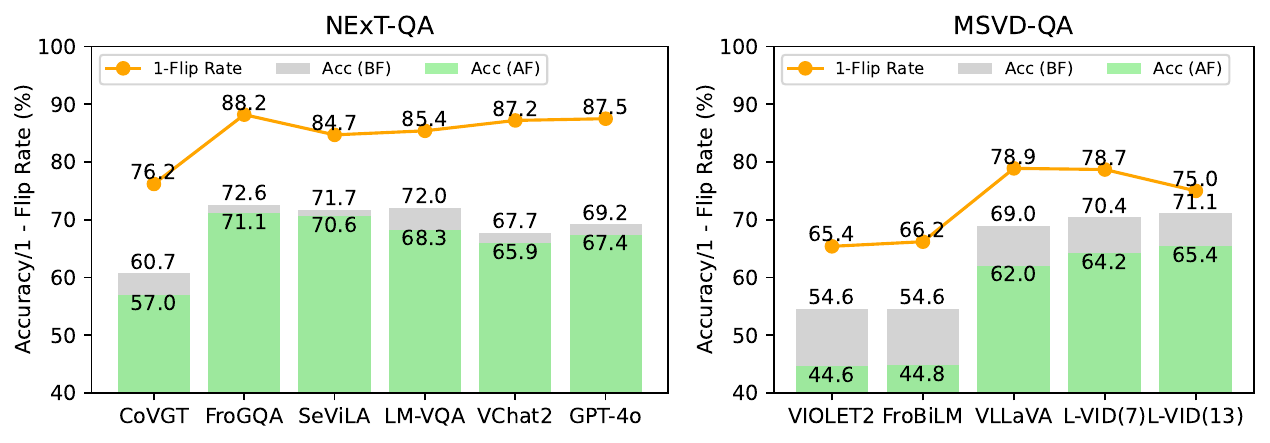}}
    \hfill
  \subfloat[Shuffle video frames. \label{fig:robust-ev-sff}]{%
        \includegraphics[width=0.49\linewidth]{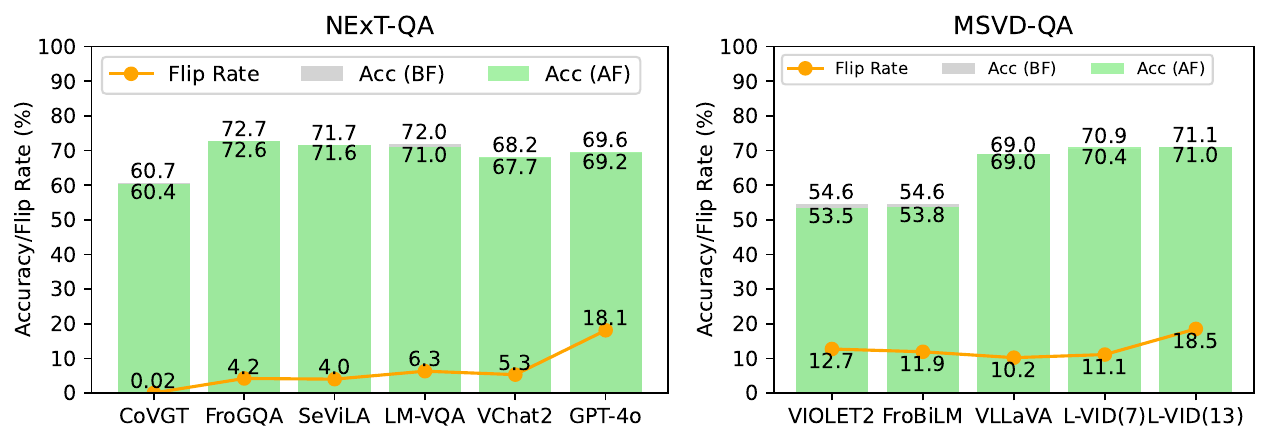}}
    \hfill
  \subfloat[Shuffle question words. \label{fig:robust-ev-sfq}]{%
        \includegraphics[width=0.49\linewidth]{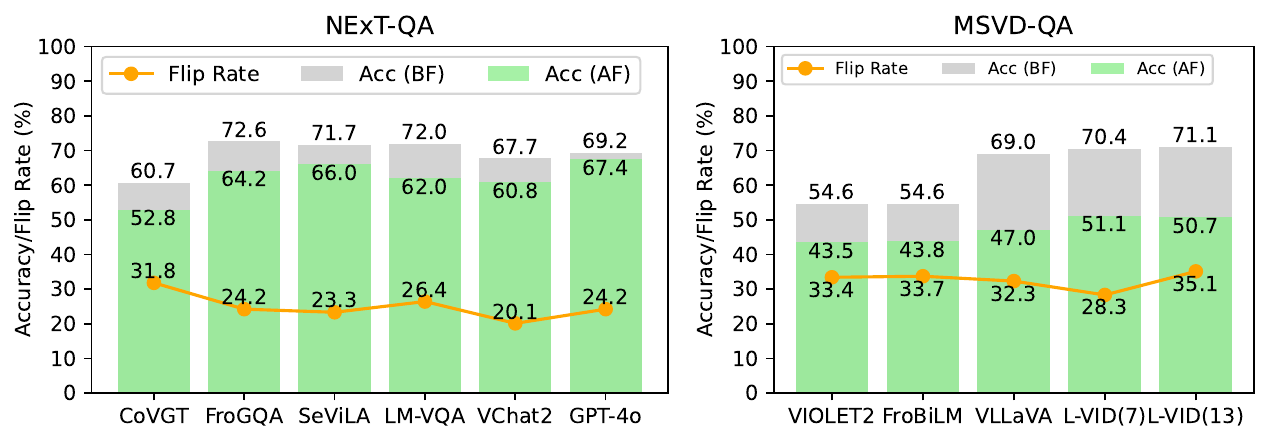}}
  \hfill
 \subfloat[Remove question keywords (question words, nouns, and verbs). \label{fig:robust-ev-rw}]{%
        \includegraphics[width=1.0\linewidth]{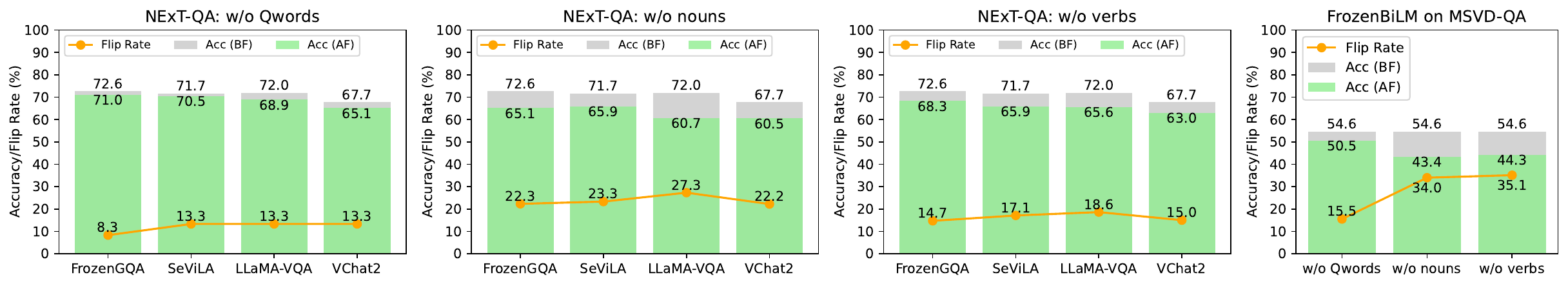}}
  \caption{Results of robustness on both MCQA (NExT-QA) and OEQA (MSVD-QA). The answers of Video-LLaVA (VLLaVA) and LLaMA-VID (L-VID, 7B, and 13B) are judged by GPT-3.5-turbo following the original papers. Also, their flip rates are calculated based on the %judgement (yes/no) 
  yes/no responses of GPT. (FroGQA: FrozenGQA. FroBiLM: FrozenBiLM. LM-VQA: LLaMA-VQA.)}
    \label{fig:robust-res}
    \vspace{-0.33cm}
\end{figure*}

For probing of model robustness in \textbf{data diversifying}, the results (both accuracy and keep rate (1 - flip rate)) in Fig.~\ref{fig:robust-iv-sp} and \ref{fig:robust-iv-rq} show that almost all models deteriorate performances, especially in OEQA. Yet, we surprisingly find that GPT-4o' accuracy gets improved by 3\% when adding spoken phrases, we attribute this strength of GPT-4o to its learning from numerous real-life user cases. Additionally, we highlight the weak behavior of the general-purpose Video-LLMs in answering questions with spoken prefixes in OEQA. For example, the accuracy degradation rates and flip dates of Video-LLaVA and LLaMA-VID are both higher than those of non-LLM method VIOLETv2. Finally, comparing the same model (LLaMA-VID) of different LLM sizes (7B and 13B) reveals that larger LLMs improve accuracy but not necessarily invariance to data diversity.

\begin{figure*}[!t]
\centering
\includegraphics[width=\linewidth]{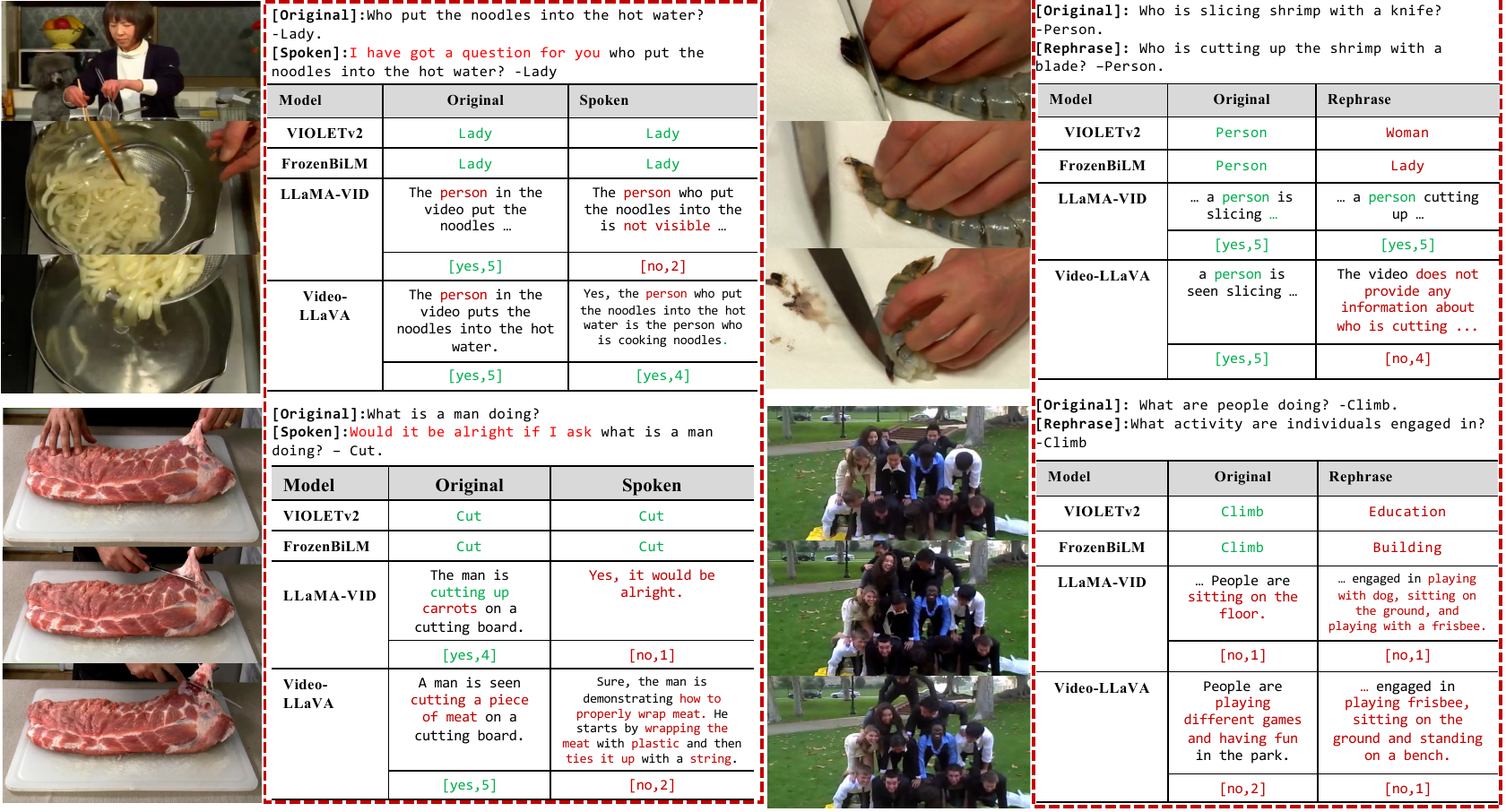}
\caption{LLaMA-VID and Video-LLaVA predictions after adding spoken phrases and rephrasing questions. The model answers are provided together with the GPT-3.5-turbo evaluation, \eg, [yes, 4] means that GPT-3.5 thinks the prediction matches the ground-truth answer with a score of $4\in\{1, \cdots, 5\}$.}
\vspace{-0.4cm}
\label{fig:rob-repq-res}
\end{figure*}

For better interpretation, we visualize some predictions in Fig.~\ref{fig:rob-repq-res}. The general purpose Video-LLMs (\eg, LLaMA-VID and Video-LLaVA)
tend to change their answers with the edited questions even though the questions' meanings remain the same. Generally, they prefer giving more specific answers, but always result in hallucinations or ignore the actual questions (\eg, refer to examples at the bottom of Fig.~\ref{fig:rob-repq-res}). A likely reason could be that the models are not instruction-tuned to answer such edited questions. Another cause could be that the modified questions are often longer and appear more complex than the original ones. Consequently, LLMs are induced for more detailed responses point by point, which increases the risk of hallucination. 
In conclusion, the experiments demonstrate that the models are not sufficiently robust to language diversifying. 

\begin{figure}[t!]
\centering
\includegraphics[width=\linewidth]{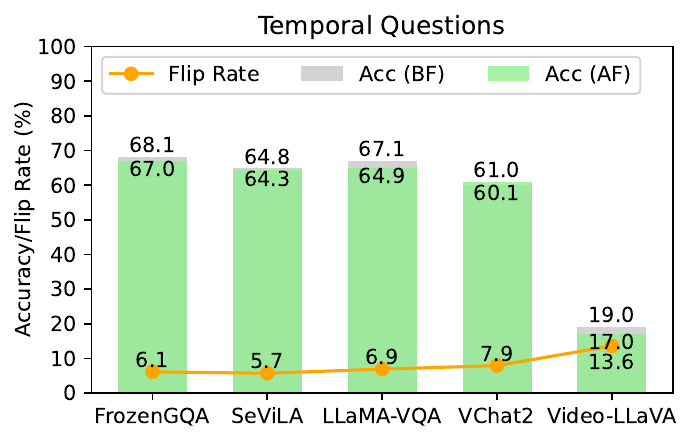}
\caption{Results in answering temporal questions after shuffling video frames. The questions are selected from ActivityNet-QA (size of $\sim$700) for Video-LLaVA in open-ended QA and NExT-QA (size of $\sim$1K) for other models in multi-choice QA.}
\label{fig:rob-ssf-temp}
\vspace{-0.4cm}
\end{figure}

For probing of model robustness in \textbf{data poisoning}, Fig.~\ref{fig:robust-ev-sff} shows that after \textbf{shuffling the video frames}, all models' accuracies have little to no change, be it on MCQA or OEQA. Also, the flip rates are small. The exception, however, 
is GPT-4o, which has % shows an exceptionally 
a $\sim$ 4 to 5$\times$ higher flip rate than other models. This result echos the findings in Sec.~\ref{sec:temp} and \ref{sec:ground} where GPT-4o conditions better on the video content for question answering. Yet, for all models, the negligible fluctuation of accuracy indicates a strong static bias in making predictions. 
To be more precise, we additionally evaluate the effect of frame shuffling on a subset of temporal questions whose answers are related to content ordering, and intuitively, should be more sensitive to frame shuffling. 
The results in Fig.~\ref{fig:rob-ssf-temp} show that shuffling frames again affects little the model performances in answering even temporal questions. This reflects that the models are stubborn and are not anchored on content order for answering temporal questions. 
Combining with our previous observation on BlindQA experiments (see Sec.~\ref{sec:ground} and Fig.~\ref{fig:ground-res}), we conclude that existing Video-LLMs do not really capture video temporality though they ground on videos for answer predictions. 
 
Compared with shuffling frames, Fig.~\ref{fig:robust-ev-sfq} shows that model performances degenerate more significantly when {\bf shuffling question words}. Yet, they still maintain 86\%$\sim$97\% of their original performances on MCQA, and 70\%$\sim$80\% on OEQA. Comparing among models, Video-LLMs generally show higher accuracy and lower flip rates than non-LLM methods. In particular, the performance of GPT-4o on NExT-QA only decreases marginally by 1.8\%. The results indicate Video-LLMs' superiority in resilience of question poisoning. Nonetheless, this kinds of resilience departs from human intuition in answering questions without syntax. 
% with shuffled question words.

Additionally, the probes of shuffling frames and questions, combined with the previous temporal probe in Sec.~\ref{sec:temp}, suggest that existing models have little to no capability of capturing content position or ordering information in neither language nor video inputs. A major reason could be the weak position embeddings.
% Current models all adopt some form of  positional embedding,
No matter the entangled absolute position embeddings as in BERT, or the decoupled relative position embeddings in other language models, such embeddings currently do not fully reflect the ordering information and point to a potential direction for further investigation.

\begin{figure*}[t!]
\centering
\subfloat[Truncate Questions. \label{fig:robust-truncate}]{%
        \includegraphics[width=0.6\linewidth]{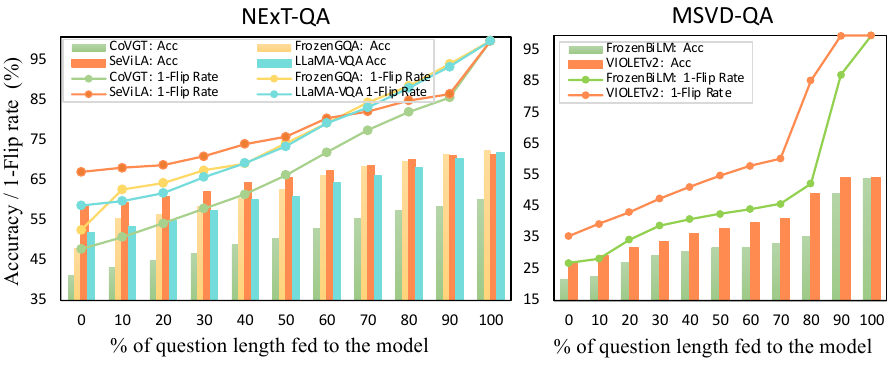}}
    \hfill
\subfloat[Long-tailed Prediction. \label{fig:fair}]{%
        \includegraphics[width=0.37\linewidth]{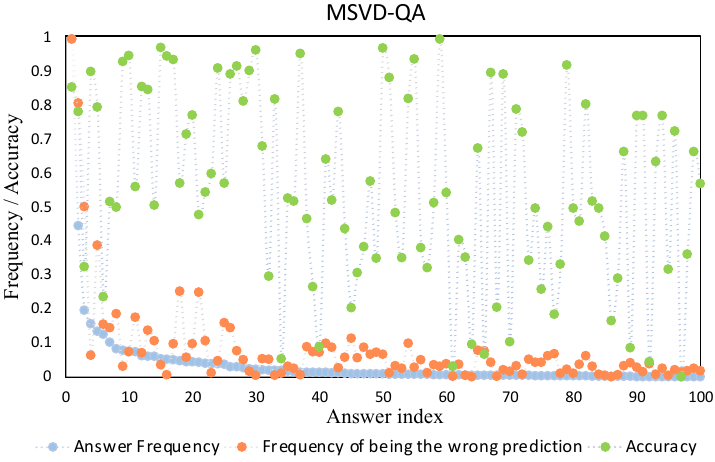}}
\caption{Model behavior with (a) truncated question inputs and (b) long-tailed answer distribution. Results with zero-length questions in (a) depict the scenario of ``\emph{knowing the answer without or ahead of the question}''. The predictions for the answers of top-100 frequency are shown in (b), where the answers are arranged in descending order according to their frequency, \eg, 0 indexes the answer of highest frequency.}
\label{fig:robfair}
% \vspace{-0.3cm}
\end{figure*}

For \textbf{removing question key words}, Fig.~\ref{fig:robust-ev-rw} shows that models suffer %relatively 
more performance loss and flip more predictions when removing nouns and verbs compared to removing the question words. This holds true for both MCQA and OEQA. The observation differs from \citep{agrawal2016analyzing} where question words are more crucial than verbs. However, this is reasonable as we analyze VideoQA (\vs image VQA), where the QAs are mostly related to actions.

Aside from removing particular words, Fig.~\ref{fig:robust-truncate} revisits the models' behavior with truncated question inputs as in ~\citep{agrawal2016analyzing}, but in the context of LLMs. Overall, Video-LLMs show stronger performance in answering incomplete question than non-LLM methods, but the specific behavior slightly varies among MCQA and OEQA. 
For better analysis and comparison, we follow \citep{agrawal2016analyzing} to highlight the models' behavior at half (50\%) input length. 
% of question inputs. 

Concretely, in MCQA, the keep rate (1 - flip rate) at this \emph{half} point indicates that the Video-LLMs %have \hl{converged} on 
predict a fixed answer for approximately 75\% (refer to SeViLA on NExT-QA) of the samples 
by listening to merely half the questions. %While 
This fixed answer ratio is higher than that of 60\% for non-LLM method (refer to CoVGT on NExT-QA), the corresponding accuracy is also higher of 60\% (\vs~ 50\%). This suggests that with incomplete question inputs, video-LLMs consistently predict %more to the 
correct answers more so than non-LLM methods.
In OEQA, the fixed answer rate of Video-LLM (FrozenBiLM) at the \emph{half} point is about 40\%, which is smaller than that of non-LLM method VIOLET2, \eg, 55\%. Yet, a joint consideration of the lower accuracy ($\sim$30\%) suggests that FrozenBiLM converges less to the wrong answers. Together with %what have observed in 
the observations of~\citep{agrawal2016analyzing} for CNN+RNN and attention models, it seems that models with better accuracy are less likely to predict wrong answers even with incomplete question inputs.

Additionally, Video-LLMs achieve higher MCQA accuracy (\eg, $\sim$60\% of SeViLA) even \emph{without} the questions. This result echoes our findings in Sec.~\ref{sec:mmreason} 
on video-answer short-cuts for decision making. Regarding OEQA, an interesting observation is that VIOLETv2 outperforms FrozenBiLM when fed with incomplete questions, but such superiority gradually disappears with the increase of question words. We suspect this is because VIOLETv2 finetunes on BERT (\vs~ frozen LLMs) with masked language modelling.  The masking mechanism allows cross-modal matching with partial and incomplete sentences, and may conflict with the underlying intent of the probe.  %corresponds to the nature of the prob.

\subsection{Generalizability}
Video-LLMs, by virtue of using LLMs, are assumed to encode some common sense and open-world knowledge \citep{kojima2022large,ko2023large}. To that end, their decisions should be less affected by the %limited 
QA training data and generalize better. Our previous probes %may have 
implicitly test model generalization to a certain extent 
by curating new data. This section studies two classic generalization scenarios more explicitly: % in this section: 
generalization to \textbf{long-tailed} data within dataset 
and generalization to \textbf{out-of-distribution} (OOD) data across datasets. %, \eg, questions of different types or involving different kinds of videos. This type of data are considered 
Both long tails and OOD data conflict and do not reflect the %violate ing a lot with the
training data on which the models are developed. It is worth mentioning that the general-purpose and tool-based Video-LLMs have already shown
excellent zero-shot performance in our previous probes. Thus, we focus on analyzing the specialized Video-LLMs that are fintuned on the target datasets in this section. 

\subsubsection{Generalization to long-tailed data}
For this experiment, we investigate the models' predictions at a per-answer basis. We consider open-ended QA with close-set classification setting where each answer is treated as a class label, \eg~MSVD-QA~\citep{xu2017video}. Specifically, we analyze an answer' frequency and its accuracy, as well as the frequency of it being a false positive prediction. An answer's accuracy stands for the proportion of correctly answered questions over all questions with that answer as ground-truth answer.
We experiment with FrozenBiLM and VIOLETv2 as representatives of Video-LLMs and non-LLMs for close-set classification respectively. 

We present the results of FrozenBiLM in Fig.~\ref{fig:fair}. VIOLETv2 shares almost the same distribution and is omitted from the plot for clarity. 
The answers' frequency distribution and accuracy distribution has a correlation coefficient of only $r=0.2$ for both FrozenBiLM and VIOLETv2.  In contrast, the coefficient value between 
answers' frequency distribution and the frequency distribution of an answer being a false positive prediction is 0.9 for both methods.
The experimental results suggest that there is little to no correlation between an answer's frequency and its accuracy for both LLM and non-LLM methods. However,
there is a strong positive correlation between an answer's frequency and the frequency of it being a false positive prediction, and this stands true for both LLM- and non-LLM based methods. 
While we suspect that the LLM in FrozenBiLM may not be large enough ($\sim$1B), the experiments indeed suggest that finetuned (or specialized) Video-LLMs are similar to non-LLM method in generalizing to long-tailed data distribution. 

\begin{figure*}[t!]
\centering
\includegraphics[width=\linewidth]{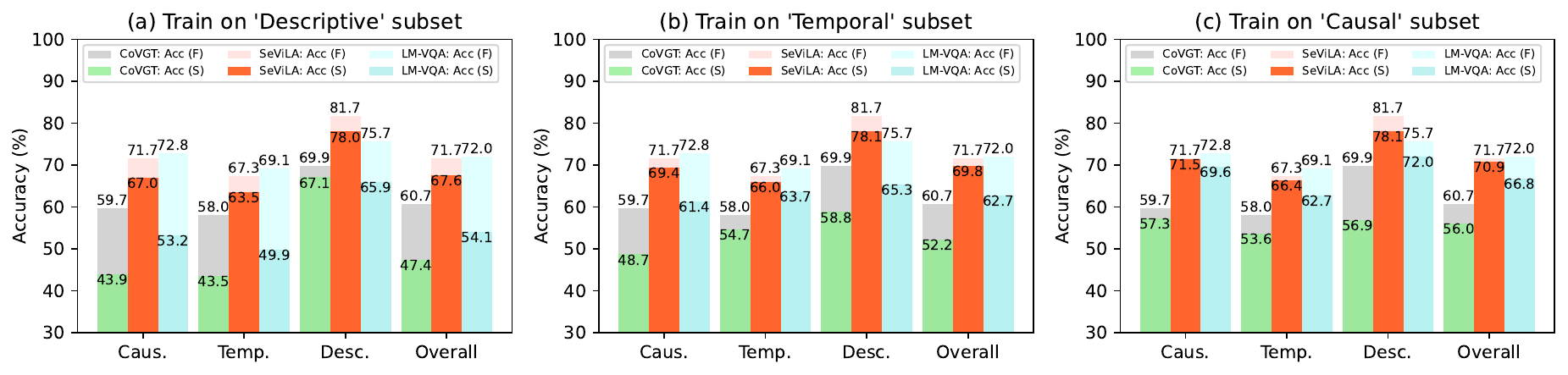}
% \vspace{-0.8cm}
\caption{Generalization across different question types. (F)/(S) denote training on the Full/Subset.}
\label{fig:generalize-res-q}
\vspace{-0.3cm}
\end{figure*}

\subsubsection{Generalization to OOD data} \label{sec:gener}
The test samples in all previous probes are either derived from the same types of questions or from the same dataset. Here, we focus on generalization across different question types and datasets.

For different question types, we experiment with NExT-QA \citep{xiao2021next} as it 
has three distinct question groups: Causal (``\texttt{why and how}"), Temporal (``\texttt{what $\cdots$ before/after/when}") and Descriptive (``\texttt{what/where is}"). We investigate models' behavior when training with data from one of the three groups and testing with samples from the other two.
For different datasets, we test the models pretrained on NExT-QA on %another 
two recent datasets: EgoSchema \citep{mangalam2023egoschema} for 1st-person view long video understanding and Perception Test \citep{puatruaucean2023perception} for fine-grained
visual perception and reasoning. These datasets are selected as they are considered one of the most prominent benchmarks for  %considered either significant for 
understanding 1st-person view visual contents and fine-grained human-object interactions, respectively.  
\begin{figure}[t!]
\centering
\includegraphics[width=1.0\linewidth]{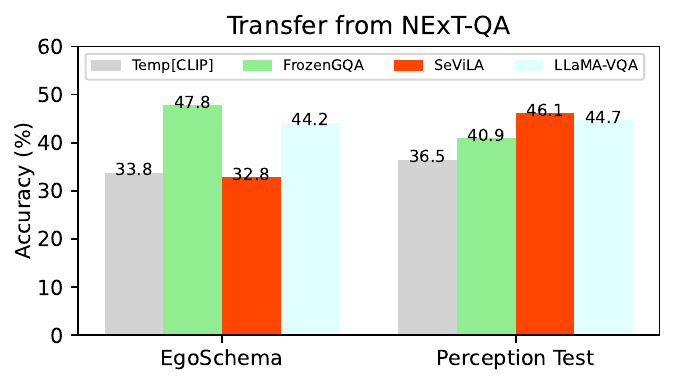}
\caption{Generalization across different datasets.}
\label{fig:generalize-res-d}
\vspace{-0.4cm}
\end{figure}

Fig.~\ref{fig:generalize-res-q} shows that Video-LLMs generally maintain stronger QA performance when transferring across different question types compared to non-LLM method (\eg, CoVGT). 
A comparison among figures of (a)/(b)/(c) suggests that training on temporal/causal questions often yields better performance than training on simple descriptive questions.
In particular, Video-LLMs (SeViLA and LLaMA-VQA) have marginal or no accuracy declines on descriptive questions when trained with temporal questions as opposed to training on the corresponding descriptive questions; performance even improves when training with causal questions. 
While a possible reason is that there are more causal and temporal questions than descriptive questions in the training data, 
we intriguingly find that the
non-LLM method contrasts with LLM method and achieves the highest accuracy of 67.1\% (\vs~ 58.8\% and 56.9\%) on descriptive questions when trained with descriptive questions.
Such discrepancy thus reflects Video-LLMs' superior behavior in generalizing to descriptive questions even if they are not trained to do so. Yet, we note that the best results for all model across different question groups are always obtained by learning on the full training set.

\begin{figure}[!t]
\centering
\includegraphics[width=\linewidth]{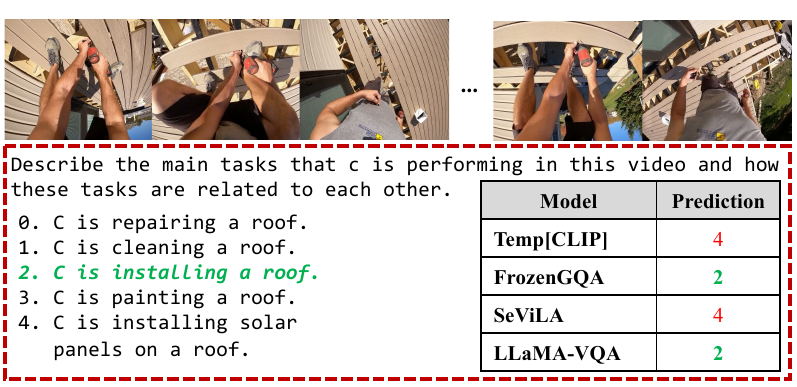}
\caption{Transfer the models pretrained on NExT-QA to EgoSchema.}
% Models that perform well on NExT-QA do not necessarily win on EgoSchema.}
\label{fig:transfer-res}
\vspace{-0.4cm}
\end{figure}
Among the models, SeViLA generalize the best among question types. However, its superiority disappears when generalizing to EgoSchema \citep{mangalam2023egoschema}. For example, its accuracy is 32.8\% which is even worse than that of non-LLM method Temp[CLIP], \eg~33.8\% (refer to Fig.~\ref{fig:generalize-res-d}). A prediction example in Fig.~\ref{fig:transfer-res} suggests that SeViLA fails to holistically understand the long video content. 
This is reasonable as SeViLA identifies only a limited number of keyframes (4 out of 32) for QA, which is a departure from 
EgoSchema which defines temporal certificate to challenges \emph{long-form} video understanding. Nevertheless, 
all Video-LLMs remain their superior performance over non-LLM method (Temp[CLIP]) in generalizing to Perception Test \citep{puatruaucean2023perception}. 
The observations, together with our experiments on NExT-OOD in Sec.~\ref{sec:mmreason} and the experiments in \citep{mangalam2023egoschema}, suggest that Video-LLMs, even the specialized or finetuned ones, commonly generalize well to OOD data. Yet this is not guaranteed for each specific model.

%% file: discussion.tex
\section{Discussion \& Future Directions}\label{sec:discussion}

\subsection{Robustness and Trustworthiness} 
All our probes showed that the open-sourced Video-LLMs perform well for standard question-answering.  However, these models are susceptible to adversarial language modifications while remain stubbornly unaffected to analogous video modifications. These findings reveal a lack of robustness, underscoring the urgent need for interpretability in Video-LLM development. There are many directions for improvement. We outline some of them, dividing them up from the perspective of data, modeling, and learning.

\textbf{From a data perspective}, a direct solution could be to apply our strategies of designing adversarial probes to augment the training data. Our simple attempt (see Appendix \ref{sec:appsolution}) shows that this can effectively improve the model performance on the probe data 
and the original test data. Aside from training data, new video testing benchmarks are emerging \citep{li2023mvbench,fu2024video,liu2024tempcompass,li2024videovista}.  Yet almost all of them target comprehensive video understanding, with little regard for interpretability. Related benchmarks such as VIP \citep{himakunthala2023let} and NExT-GQA \citep{xiao2023can} are thus valuable efforts in developing trustworthy Video-LLM techniques and deserve more research attention.

\textbf{From a modeling perspective}, VideoQA models are imbalanced in terms of their language and visual representations. The language encoders are drawn from ever-larger and more powerful language models which dominate over comparatively weaker visual representations, where  CLIP~\cite{radford2021learning} and its variants~\cite{ilharco_gabriel_2021_5143773,sun2023eva} 
remain de-facto standards even though they struggle to represent fine-grained details. More importantly, from an image to a video level, the temporal and dynamic modeling tends to be primitive. Hence, exploring and improving the image or video encoders would contribute to more robust models. 

Finally, \textbf{from a learning perspective}, existing models learn to directly output the final answers in an auto-regressive manner. Directly targeting the answers (especially lengthy ones) 
enables the models to ignore the visual contents and take whatever shortcuts they can (\eg, hallucinate based on commonsense knowledge from LLMs) to make plausible but often incorrect responses (Sec.~\ref{sec:robust}). As such, we believe the focus on intermediate reasoning steps and post-verifications can benefit model developing towards robustness and trustworthiness.
Some recent studies in multimodal chain-of-thought \citep{wei2022chain,zhang2023multimodal,fei2024video} and modular multi-agents \citep{min2024morevqa,shang2024traveler} provide insights in this direction and show improved robustness and interpretability. 

\subsection{Efficient and Fine-Grained Long-Range Understanding}
GPT-4o's superior behavior in VideoQA (see  Fig.~\ref{fig:intro} and \ref{fig:ground-res}), as well as the ascent of SOTA performance on 
benchmarks like MSVD-QA and MSRVTT-QA suggests that coarse and holistic understanding of short videos (less than 20s) are solvable by scaling up existing Video-LLMs or prompting powerful image MLLMs (like GPT-4o). Yet the understanding of fine-grained details \citep{xiao2021next} or long-range videos \citep{mangalam2023egoschema,fu2024video} remains out of reach.  These videos offer additional challenges that require more dedicated analysis and architecture design. 

One such hurdle comes from the \emph{`needle in the haystack'} problem - %which stands for 
the search in the long video for short moments (or keyframes) that carry the correct answer. Existing benchmarks such as NExT-GQA \citep{xiao2023can} and QAEgo4D \citep{datta2022episodic} embody such challenges, yet effective techniques are lacking. Recent solutions (\eg, MoReVQA \citep{min2024morevqa}, VideoAgent \citep{wang2024videoagent,fan2024videoagent} and TraveLER \citep{shang2024traveler}) show promise. However, they are relatively inefficient, in that they traverse the entire video frame-by-frame, or segment-by-segment, and perform LLM inference multiple times. Efficient understanding of such long videos is gaining attention but continues to be an open problem for the foreseeable future.

The second challenge is a fine-grained understanding of long-range videos. The concept of large temporal certificates, as defined in EgoSchema \citep{mangalam2023egoschema}, highlights accurately differentiating the visually very similar long activities (\eg, teaching someone to play \vs~play with someone). Our studies in Sec.~\ref{sec:gener} show that existing Video-LLMs targeting short clips do not generalize well to such videos. Traversal-based solutions are ineffective as the task requires reasoning over and weaving together (potentially distant) frames or segments to achieve holistic understanding. Recent memory- and caption-based solutions (\eg, MoReVQA \citep{min2024morevqa}, LLoVi \citep{zhang2023simple} and VideoAgent \citep{fan2024videoagent}) as well as the direct long-range encoding solutions (\eg, LongVA \citep{zhang2024long}) offer some insights.  However, fine-grained understanding, especially in an efficient manner, remains a key challenge. 

\subsection{Egocentric and Embodied QA}
We primarily analyzed VideoQA with exo-centric datasets - videos from the third-person view. It should be noted that an important real-world application for VideoQA is geared towards reliable embodied assistance \citep{grauman2022ego4d,majumdar2024openeqa}. In this regard, ego-centric, or first-person view videos and associated question-answer pairs 
are of great interest and relevance. Sec.~\ref{sec:gener} shows that Video-LLMs pre-trained on third-person view VideoQA datasets
can generalize to these application scenarios.  However, their performance is weak compared with the tool-based Video-LLMs \citep{zhang2023simple} that directly use ego-centric pretrained visual descriptions \citep{zhao2023learning}, and well-prompted general-purpose MLLMs (\eg~GPT-4o) \citep{liu2024coarse}. 
Therefore, the efforts of leveraging existing MLLMs for ego-centric VideoQA will be of critical importance in initiating successful embodied assistance applications.

%% file: conclusion.tex
\section{Conclusion}\label{sec:conclusion}
This paper has vetted Video-LLMs' performance in VideoQA by probing their success and failure modes with well-designed adversarial tests. 
While Video-LLMs generally show better QA accuracy, the decrease rates and flip rates align with or even higher than non-LLM methods when faced with adversarial challenges. 
Specifically, Video-LLMs show significant limitation in coping with video temporality, of both reasoning the chronological content order and grounding the temporal moments to substantiate the answers. Also, they are unresponsive towards video perturbation while being susceptible to simple language variations of questions and candidate answers. Additionally, Video-LLMs, after finetuned, are not necessarily generalize better. 
Understanding these limitations is crucial for developing future Video-LLMs and VideoQA techniques, where we also conclude some promising directions to proceed.

%% file: appendix.tex
% We first introduces our data curation strategy in Sec.~\ref{sec:apptemp}, \ref{sec:appmm} and \ref{sec:approbust}. Then, we detail our approach of adopting GPT-4o for VideoQA in Sec.~\ref{sec:appgpt4o}. Finally, we investigate data augmentation with our probing strategies in Sec.~\ref{sec:appsolution}. 
\subsection{Temporal Probes} \label{sec:apptemp}
We adopt different methods to generate the \textbf{Temporal Exchange} and \textbf{Temporal Description} data described in Sec.~\ref{sec:temp}.  Yet, both are followed by human (the first 4 authors) checking and correction to ensure the quality. 

Specifically, to get the data of \textbf{Temporal Exchange}, we prompt GPT-4 by giving the original question and correct answer pairs. In our implementation, we find it is better to use different prompts to handle questions about ``before/after'' and questions about ``when/while''. Specific prompts are presented in Tab.~\ref{tab:instructions_temp}. We find the generated QAs are quite good and only do small effort in correction.

To get the data of \textbf{Temporal Description}, we directly parse the syntactic structure of the questions according to the time signal words ``before/after/when/while". Specifically, we keep the question part ahead of the time words. For instance, we derive the new question ``\texttt{what did the person do?}" from the original question ``\texttt{what did the person do after she took the black item away?}'' . To curate the options, we use different time words to combine the two terms ``\texttt{she took the black item away}" and ``\texttt{pat animal}", and ensure only one combination is correct regarding the content order of the video. Also, we keep the position of the correct answer unchanged. Additionally, we find there are temporal location descriptions for some questions, such as ``\texttt{ ... in the middle of the video}". To handle this, we trim off such description and append it behind the new question. Note that the grammatical issues such as ``\texttt{before pat animal}" \vs~ ``\texttt{before patting animal}" are ignored, since them do not affect the overall meanings of the questions and answers.
\vspace{-0.3cm}

\subsection{Multi-Choice Short-Cuts} \label{sec:appmm}
To test the short-cuts in candidate answers, we obtain the edited Question-Answer (QA) and Video-Answer (VA) data by prompting GPT-4, again followed by human checking and correction. Our specific prompts are listed in Tab.~\ref{tab:instructions_mm}. For human checking, we specifically check the edited negative options to ensure that they are actual wrong answers with regard to the question and video contents. 
\vspace{-0.3cm}

\subsection{Robustness} \label{sec:approbust}
To obtain questions with \textbf{spoken} prefixes, we first prompt GPT-4 to generate a set of spoken phrases that humans typically use before asking questions as shown in Tab.~\ref{tab:appspoken}. Then, we randomly select from these spoken phrases and prepend them to the questions. To obtain the \textbf{rephrased} questions, we simply prompt GPT-4 using the instructions specified at the bottom of Tab.~\ref{tab:instructions_mm}. We find the generated questions are quite good and only do small correction.
\vspace{-0.3cm}

\subsection{GPT-4o for VideoQA} \label{sec:appgpt4o}
Following popular methods, we first decode each video at 3 frames per second (fps) and then uniformly sample 32 video frames. We then feed all of the sampled 32 frames together with the question and answer into GPT-4o and prompt it to answer the question based on the sequence of image contents. Specific prompts are shown in Tab.~\ref{tab:appgpt4o}. Particularly, for the PosVQA experiment described in Sec.~\ref{sec:ground}, we sample all frames (after decoding at 3 fps) from the positive temporal moments and uniformly sample 32 frames if the total frames of a segment is large than 32. This is because we find that, for a non-trivial number of samples, only a single frame is positive if we uniformly sample 32 frames from the whole video. Finally, in all experiments, we only experiment with 40\% of the original data considering the time and API expenses. Also, we choose not to perform GPT-4o on open-ended QA for answer evaluation issues.
\begin{table}[t!]
\centering
\begin{minipage}{0.9\columnwidth}\vspace{0mm}    
    \centering
    \caption{Examples of spoken prefixes.}
    \begin{tcolorbox} 
        \centering
        \hspace{-6mm}
        \begin{tabular}{p{0.8\columnwidth}}
        \hspace{1mm}
        \begin{minipage}{1.0\columnwidth}
        Can I ask you a quick question \\
        Would you mind if I asked \\
        Could I pick your brain for a second? \\
        I hope you don't mind me asking, but \\ 
        I've been meaning to ask you \\
        Do you have a moment to answer a question? \\
        Can I get your answer on \\
        Would you be able to answer \\
        I need your answer on \\ 
        I'm trying to find out \\
        May I ask for your answer on \\
        Could you help me understand \\
        Is it okay if I ask \\
        I'm curious about \\
        Would it be alright if I ask \\ 
        I've got a question for you. \\
        Could you clarify something for me? \\
        Do you know \\
        I want to know \\
        Can you tell me \\
        Would you happen to know \\
        May I know \\
        Could you please let me know \\
        Could you please tell me \\
        ... 
        \end{minipage}
        \end{tabular}
    \end{tcolorbox}
\label{tab:appspoken}
\end{minipage}
\end{table}

\setlength{\tabcolsep}{15pt}
\begin{table}[t]
	\center
	\small
	\caption{Behavior of TD data augmentation.} 
        % \vspace{-0.3cm}
	\label{tab:solution-td-aug}
	\scalebox{0.8}{
		\begin{tabular}{l|c|c|ccc}
			\hline\hline
			Methods & TO & TE  & TD & \\ \hline
			FrozenGQA & 67.7 & 46.1 & 32.3\\
                \hline
 			FrozenGQA (TD Aug.) & 68.3 & 52.0 & 92.5 \\  
            \hline
		\end{tabular}
	}
	\vspace{-0.4cm}
\end{table}
\subsection{Effectiveness of Data Augmentation} \label{sec:appsolution}
For improvement, we take the Temporal Description (TD) probe (introduced in Sec.~\ref{sec:temp}) as an example to explore the effects of data augmentation, given that all models hardly survive on this probe. In our implementation, we curate new TD data based on the original \emph{temporal} subset of training data, mirroring how we curate TD test data. We then add the curated TD data into the original training data of NExT-QA for training ($\sim4.5$K). The results in Tab.~\ref{tab:solution-td-aug} show that this simple data augmentation significantly boosts the model performance on both the temporal description (TD) and temporal exchange (TE) probes, and slightly improves the performance on the original test set (TO) for temporal understanding. The consistent improvements on all test settings thus demonstrate the effectiveness of such data argumentation towards more robust temporal understanding. 
Additionally, we observe profound improvements on TD test probe. We speculate that the curated TD test data oriented for zero-shot testing may leak statistical bias when training with such forms of data.

\begin{table*}[t!]\centering
\begin{minipage}{1.0\linewidth}\vspace{0mm}
\centering
\caption{The list of instructions for prompting GPT-4 to produce the temporal probing data. Note that the QA examples in the prompts may vary a little bit to cope with all samples.}
\begin{tcolorbox} 
    \centering
    \small
     \hspace{-6mm}
\begin{enumerate}
\setlength{\itemsep}{2pt}
    \item \textbf{Temporal Exchange (before/after):} ``\texttt{
    You are an excellent video question generator. Please help to change the before (after) questions into after (before) ones. \\
    Examples: \\
    Input: "what did the lady do before wiping the baby's mouth? Take napkin." \\
    Output: "what did the lady do after taking the napkin? Wipe the baby's mouth." \\
    Input: "What did the man in grey do before the plane took off? Hold the plan." \\
    Output: "What happened to the plane after the man holds the plane? Plane take off." \\
    Input: "what was the boy in yellow doing before the boy in blue turned his head behind at the middle? Pick up something from ground." \\
    Output: "What was the boy in blue doing after the boy in yellow picks something up from the ground at the middle? Turn his head behind." \\
    Input: "what did the lady in blue do when she reached the stairs? Sit on stairs." \\
    Output: "What did the lady in blue do before she sits on stairs? Reach the stairs." \\
    I will give you the original question-answer pairs, please only return the generated QA pair. The answer should start after the '?'.
    }''
    \item \textbf{Temporal Exchange (when/while):} ``\texttt{
    You are an excellent video question generator. Please help to change the temporal question-answer pairs into different ones. For example, "what does A do when/while B ... " should be changed to "what does (happened to) B do when/while A ...". Make sure the generated question-answer pair share the same video content with the intput question-answer pair. The generated questions should be natural and meaningful. \\
    Examples: \\
    Input: "what did the girl do when the lady was patting the dog at the start? Walk across." \\
    Output: "what did the lady do when the girl walked across at the start? Pat the dog." \\
    Input: "What does the man in blue do when the car titles at a sharp angle near the end? stand up." \\
    Output: "what happened to the car when the man in blue stand up near the end? Car titles at a sharp angle." \\
    Input: "what happened to the man in grey when the man in blue socks jump onto the man in grey? carry him." \\
    Output: "What did the man in blue do before the man in grey carry him? Jump onto the man in grey." \\
    Input: "what happened to the man in grey when the man in blue socks jump onto the man in grey? carry him." \\
    Output: "What did the man in blue do before the man in grey carry him? Jump onto the man in grey." \\
    I will give you the original question-answer pairs, please generate and only return the QA pair. The answer should start after the '?'
    }''
\end{enumerate}
\end{tcolorbox}
    \label{tab:instructions_temp}
\end{minipage}
\end{table*}

\begin{table*}[t!]\centering
\begin{minipage}{1.0\linewidth}\vspace{0mm}
\centering
\caption{The list of instructions for prompting GPT-4 to finding and mitigating QA and VA short-cuts, as well as for rephrasing questions.}
\begin{tcolorbox} 
    \centering
    \small
     \hspace{-6mm}
\begin{enumerate}
\setlength{\itemsep}{2pt}
    \item \textbf{QA short-cuts:} ``\texttt{You are an excellent video question generator. I need you help to amend the multiple-choice answers for the question. All candidate answers should contain the key words from the question. The edited answers should be succinct as the original ones, with each adhering to 2 to 6 words. For example:  the original question is `What did the person do after the animal started sniffing the black item on the bed?' and the options are `[jump on it, changing diaper, remove black item, play with toy, moves away]'. The refined options should be: [jump on bed, changing diaper for animal, remove black item, play with black toy, moves away from bed]. I will give you original questions and corresponding options. please return the refined options only. If most of the original options contain the key words from the question, then please return am empty list `[]'.}''
    \item \textbf{VA short-cuts:} ``\texttt{You are an excellent video question generator. I need your help to amend the multiple-choice answers for the question. Do small edit to ensure each answer in the options correlates to the video labels but differ from each other. The edited answers should be succinct as the original ones, with each adhering to 1 to 6 words and keep the same order. For example, the video labels are: [gift, doorway, doll, girl, boy, video game, stool, grandmother, treat, couch, floor, picture frame, tree, christmas, pink, christmas tree, woman, toy, chair, walk, mother]. The original question is: `why did the boy pick up one present from the group of them and move to the sofa', and the options are: [share with girl, approach lady sitting there, unwrap it, playing with toy train, gesture something]. Example of well-refined options is: [share with girl, approach lady sitting on sofa, unwrap gift, playing with toy train, gesture towards woman]. If the answer already relates to the video labels, keep it unchanged. I will give you the original questions, corresponding answer options and video labels. Please only return the amended options without other context}''
    \item \textbf{Question Rephrase:} ``\texttt{You are an excellent video question generator. I need your assistance to rephrase the given question to another one with the same meaning but with different language format. For instance, if the original question is `why did the baby cry?', well-generated questions could be `what caused the infant to weep?' or `Why is the kid crying'. Please return 1 best rephrased question for each question I give to you.}''
\end{enumerate}
\end{tcolorbox}
    \label{tab:instructions_mm}
\end{minipage}
\end{table*}

\begin{table*}[t!]
\centering
\begin{minipage}{1.0\linewidth}\vspace{0mm}    
    \centering
    \caption{Prompts for GPT-4o to perform VideoQA. For different experiments, we will replace the example inputs and outputs with the respective QAs in that experiments. In particular, for visual grounding experiment, we find that the highlighted sentence is very important to achieve better QA performance, regardless of whether performing grounding.}
    \begin{tcolorbox} 
        \centering
        \hspace{-6mm}
        \begin{tabular}{p{1.0\linewidth}}
        \hspace{1mm}
        \begin{minipage}{0.96\linewidth}
        ``\texttt{You are excellent at answering video questions. Task: \\
For an input question and 5 optional answers, you need to output the key (chosen from 0 to 4)
of the correct answer based on the sequence of image contents. \textbf{You also need to output an index interval between 0 to 31 (or the total number of images) 
, which indicates the images you based on to get the correct answer.}
The image sequence are sampled video frames and are sorted in time order. \\
Examples: \\
Input: why does the lady wipe the baby's mouth? "0": "baby spits out milk.", "1":"choolate stains.", "2":"strawberry stains.", "3": "there was paint on his face.", "4":"there was water spills." \\
Output: {"answer": 0, "images": [14,15]} \\
Input: how did the baby move up the stairs near the end of the video? "0": "lady carried her.", "1":"walk up.", "2":"ran up.", "3": "man hold her up.", "4":"crawl up." \\
Output: {"answer": 4, "images": [15,31]} \\
Input: What did the lady do after wipping the baby's mouth? "0": "Touch the baby again.", "1":"Turn away.", "2":"Adjust the baby in grey.", "3": "Smile.", "4":"Take napkin." \\
Output: {"answer": 4, "images": [15,15]} \\
- I will give you question and options. **Only output the correct answer key and index interval.** \\
Input: }''
        \end{minipage}
        \end{tabular}
    \end{tcolorbox}
\label{tab:appgpt4o}
\end{minipage}
\end{table*}

%% file: main.bbl
\begin{thebibliography}{97}
\providecommand{\natexlab}[1]{#1}
\providecommand{\url}[1]{{#1}}
\providecommand{\urlprefix}{URL }
\expandafter\ifx\csname urlstyle\endcsname\relax
  \providecommand{\doi}[1]{DOI~\discretionary{}{}{}#1}\else
  \providecommand{\doi}{DOI~\discretionary{}{}{}\begingroup \urlstyle{rm}\Url}\fi
\providecommand{\eprint}[2][]{\url{#2}}

\bibitem[{Agrawal et~al.(2016)Agrawal, Batra, and Parikh}]{agrawal2016analyzing}
Agrawal A, Batra D, Parikh D (2016) Analyzing the behavior of visual question answering models. In: Conference on Empirical Methods in Natural Language Processing (EMNLP), pp 1955--1960

\bibitem[{Alayrac et~al.(2022)Alayrac, Donahue, Luc, Miech, Barr, Hasson, Lenc, Mensch, Millican, Reynolds et~al.}]{alayrac2022flamingo}
Alayrac JB, Donahue J, Luc P, Miech A, Barr I, Hasson Y, Lenc K, Mensch A, Millican K, Reynolds M, et~al. (2022) Flamingo: a visual language model for few-shot learning. Advances in Neural Information Processing Systems (NeurIPS) 35:23716--23736

\bibitem[{Antol et~al.(2015)Antol, Agrawal, Lu, Mitchell, Batra, Zitnick, and Parikh}]{antol2015vqa}
Antol S, Agrawal A, Lu J, Mitchell M, Batra D, Zitnick CL, Parikh D (2015) Vqa: Visual question answering. In: Proceedings of the IEEE international conference on computer vision (ICCV), pp 2425--2433

\bibitem[{Bagad et~al.(2023)Bagad, Tapaswi, and Snoek}]{bagad2023test}
Bagad P, Tapaswi M, Snoek CG (2023) Test of time: Instilling video-language models with a sense of time. In: Proceedings of the IEEE/CVF Conference on Computer Vision and Pattern Recognition (CVPR), pp 2503--2516

\bibitem[{Bai et~al.(2024)Bai, Wang, Xiao, He, Han, Zhang, and Shou}]{bai2024hallucination}
Bai Z, Wang P, Xiao T, He T, Han Z, Zhang Z, Shou MZ (2024) Hallucination of multimodal large language models: A survey. arXiv preprint arXiv:240418930

\bibitem[{Buch et~al.(2022)Buch, Eyzaguirre, Gaidon, Wu, Fei-Fei, and Niebles}]{buch2022revisiting}
Buch S, Eyzaguirre C, Gaidon A, Wu J, Fei-Fei L, Niebles JC (2022) Revisiting the "video" in video-language understanding. In: Proceedings of the IEEE/CVF conference on computer vision and pattern recognition (CVPR), pp 2917--2927

\bibitem[{Chen et~al.(2023)Chen, Djolonga, Padlewski, Mustafa, Changpinyo, Wu, Ruiz, Goodman, Wang, Tay et~al.}]{chen2023pali}
Chen X, Djolonga J, Padlewski P, Mustafa B, Changpinyo S, Wu J, Ruiz CR, Goodman S, Wang X, Tay Y, et~al. (2023) Pali-x: On scaling up a multilingual vision and language model. arXiv preprint arXiv:230518565

\bibitem[{Chiang et~al.(2023)Chiang, Li, Lin, Sheng, Wu, Zhang, Zheng, Zhuang, Zhuang, Gonzalez, Stoica, and Xing}]{vicuna2023}
Chiang WL, Li Z, Lin Z, Sheng Y, Wu Z, Zhang H, Zheng L, Zhuang S, Zhuang Y, Gonzalez JE, Stoica I, Xing EP (2023) Vicuna: An open-source chatbot impressing gpt-4 with 90\%* chatgpt quality. \urlprefix\url{https://lmsys.org/blog/2023-03-30-vicuna/}

\bibitem[{Chung et~al.(2022)Chung, Hou, Longpre, Zoph, Tay, Fedus, Li, Wang, Dehghani, Brahma et~al.}]{chung2022scaling}
Chung HW, Hou L, Longpre S, Zoph B, Tay Y, Fedus W, Li Y, Wang X, Dehghani M, Brahma S, et~al. (2022) Scaling instruction-finetuned language models. arXiv preprint arXiv:221011416

\bibitem[{Dai et~al.(2023)Dai, Li, Li, Tiong, Zhao, Wang, Li, Fung, and Hoi}]{dai2023instructblip}
Dai W, Li J, Li D, Tiong AMH, Zhao J, Wang W, Li B, Fung P, Hoi S (2023) Instructblip: towards general-purpose vision-language models with instruction tuning. In: Proceedings of the 37th International Conference on Neural Information Processing Systems (NeurIPS), pp 49250--49267

\bibitem[{Datta et~al.(2022)Datta, Dharur, Cartillier, Desai, Khanna, Batra, and Parikh}]{datta2022episodic}
Datta S, Dharur S, Cartillier V, Desai R, Khanna M, Batra D, Parikh D (2022) Episodic memory question answering. In: Proceedings of the IEEE/CVF Conference on Computer Vision and Pattern Recognition (CVPR), pp 19119--19128

\bibitem[{Devlin et~al.(2018)Devlin, Chang, Lee, and Toutanova}]{devlin2018bert}
Devlin J, Chang MW, Lee K, Toutanova K (2018) Bert: Pre-training of deep bidirectional transformers for language understanding. arXiv preprint arXiv:181004805

\bibitem[{Dosovitskiy et~al.(2021)Dosovitskiy, Beyer, Kolesnikov, Weissenborn, Zhai, Unterthiner, Dehghani, Minderer, Heigold, Gelly et~al.}]{dosovitskiy2020image}
Dosovitskiy A, Beyer L, Kolesnikov A, Weissenborn D, Zhai X, Unterthiner T, Dehghani M, Minderer M, Heigold G, Gelly S, et~al. (2021) An image is worth 16x16 words: Transformers for image recognition at scale. In: International Conference on Learning Representations (ICLR)

\bibitem[{Dubey et~al.(2024)Dubey, Jauhri, Pandey, Kadian, Al-Dahle, Letman, Mathur, Schelten, Yang, Fan et~al.}]{dubey2024llama}
Dubey A, Jauhri A, Pandey A, Kadian A, Al-Dahle A, Letman A, Mathur A, Schelten A, Yang A, Fan A, et~al. (2024) The llama 3 herd of models. arXiv preprint arXiv:240721783

\bibitem[{Fan et~al.(2024)Fan, Ma, Wu, Du, Li, Gao, and Li}]{fan2024videoagent}
Fan Y, Ma X, Wu R, Du Y, Li J, Gao Z, Li Q (2024) Videoagent: A memory-augmented multimodal agent for video understanding. European Conference on Computer Vision (ECCV)

\bibitem[{Fei et~al.(2024)Fei, Wu, Ji, Zhang, Zhang, Lee, and Hsu}]{fei2024video}
Fei H, Wu S, Ji W, Zhang H, Zhang M, Lee ML, Hsu W (2024) Video-of-thought: Step-by-step video reasoning from perception to cognition. In: Forty-first International Conference on Machine Learning (ICML)

\bibitem[{Fu et~al.(2024)Fu, Dai, Luo, Li, Ren, Zhang, Wang, Zhou, Shen, Zhang et~al.}]{fu2024video}
Fu C, Dai Y, Luo Y, Li L, Ren S, Zhang R, Wang Z, Zhou C, Shen Y, Zhang M, et~al. (2024) Video-mme: The first-ever comprehensive evaluation benchmark of multi-modal llms in video analysis. arXiv preprint arXiv:240521075

\bibitem[{Fu et~al.(2021)Fu, Li, Gan, Lin, Wang, Wang, and Liu}]{fu2021violet}
Fu TJ, Li L, Gan Z, Lin K, Wang WY, Wang L, Liu Z (2021) Violet: End-to-end video-language transformers with masked visual-token modeling. arXiv preprint arXiv:211112681

\bibitem[{Fu et~al.(2023)Fu, Li, Gan, Lin, Wang, Wang, and Liu}]{fu2023empirical}
Fu TJ, Li L, Gan Z, Lin K, Wang WY, Wang L, Liu Z (2023) An empirical study of end-to-end video-language transformers with masked visual modeling. In: Proceedings of the IEEE/CVF Conference on Computer Vision and Pattern Recognition (CVPR), pp 22898--22909

\bibitem[{Goyal et~al.(2017)Goyal, Khot, Summers-Stay, Batra, and Parikh}]{goyal2017making}
Goyal Y, Khot T, Summers-Stay D, Batra D, Parikh D (2017) Making the v in vqa matter: Elevating the role of image understanding in visual question answering. In: IEEE Conference on Computer Vision and Pattern Recognition (CVPR), pp 6904--6913

\bibitem[{Grauman et~al.(2022)Grauman, Westbury, Byrne, Chavis, Furnari, Girdhar, Hamburger, Jiang, Liu, Liu et~al.}]{grauman2022ego4d}
Grauman K, Westbury A, Byrne E, Chavis Z, Furnari A, Girdhar R, Hamburger J, Jiang H, Liu M, Liu X, et~al. (2022) Ego4d: Around the world in 3,000 hours of egocentric video. In: Proceedings of the IEEE/CVF Conference on Computer Vision and Pattern Recognition, pp 18995--19012

\bibitem[{He et~al.(2016)He, Zhang, Ren, and Sun}]{he2016deep}
He K, Zhang X, Ren S, Sun J (2016) Deep residual learning for image recognition. In: Proceedings of the IEEE Conference on Computer Vision and Pattern Recognition (CVPR), pp 770--778

\bibitem[{He et~al.(2020)He, Liu, Gao, and Chen}]{he2020deberta}
He P, Liu X, Gao J, Chen W (2020) Deberta: Decoding-enhanced bert with disentangled attention. arXiv preprint arXiv:200603654

\bibitem[{Himakunthala et~al.(2023)Himakunthala, Ouyang, Rose, He, Mei, Lu, Sonar, Saxon, and Wang}]{himakunthala2023let}
Himakunthala V, Ouyang A, Rose D, He R, Mei A, Lu Y, Sonar C, Saxon M, Wang W (2023) Let’s think frame by frame with vip: A video infilling and prediction dataset for evaluating video chain-of-thought. In: Proceedings of the 2023 Conference on Empirical Methods in Natural Language Processing (EMNLP), pp 204--219

\bibitem[{Hochreiter and Schmidhuber(1997)}]{hochreiter1997long}
Hochreiter S, Schmidhuber J (1997) Long short-term memory. Neural computation 9(8):1735--1780

\bibitem[{Hu et~al.(2021)Hu, Shen, Wallis, Allen-Zhu, Li, Wang, Wang, and Chen}]{hu2021lora}
Hu EJ, Shen Y, Wallis P, Allen-Zhu Z, Li Y, Wang S, Wang L, Chen W (2021) Lora: Low-rank adaptation of large language models. arXiv preprint arXiv:210609685

\bibitem[{Ilharco et~al.(2021)Ilharco, Wortsman, Wightman, Gordon, Carlini, Taori, Dave, Shankar, Namkoong, Miller, Hajishirzi, Farhadi, and Schmidt}]{ilharco_gabriel_2021_5143773}
Ilharco G, Wortsman M, Wightman R, Gordon C, Carlini N, Taori R, Dave A, Shankar V, Namkoong H, Miller J, Hajishirzi H, Farhadi A, Schmidt L (2021) Openclip. \doi{10.5281/zenodo.5143773}, \urlprefix\url{https://doi.org/10.5281/zenodo.5143773}, if you use this software, please cite it as below.

\bibitem[{Jang et~al.(2017)Jang, Song, Yu, Kim, and Kim}]{jang2017tgif}
Jang Y, Song Y, Yu Y, Kim Y, Kim G (2017) Tgif-qa: Toward spatio-temporal reasoning in visual question answering. In: Proceedings of the IEEE Conference on Computer Vision and Pattern Recognition (CVPR), pp 2758--2766

\bibitem[{Jang et~al.(2019)Jang, Song, Kim, Yu, Kim, and Kim}]{jang2019video}
Jang Y, Song Y, Kim CD, Yu Y, Kim Y, Kim G (2019) Video question answering with spatio-temporal reasoning. International Journal of Computer Vision (IJCV) 127:1385--1412

\bibitem[{Kervadec et~al.(2021)Kervadec, Antipov, Baccouche, and Wolf}]{kervadec2021roses}
Kervadec C, Antipov G, Baccouche M, Wolf C (2021) Roses are red, violets are blue... but should vqa expect them to? In: Proceedings of the IEEE/CVF Conference on Computer Vision and Pattern Recognition (CVPR), pp 2776--2785

\bibitem[{Kim et~al.(2024)Kim, Choi, Lee, and Rhee}]{kim2024image}
Kim W, Choi C, Lee W, Rhee W (2024) An image grid can be worth a video: Zero-shot video question answering using a vlm. arXiv preprint arXiv:240318406

\bibitem[{Ko et~al.(2023)Ko, Lee, Kang, Roh, and Kim}]{ko2023large}
Ko D, Lee J, Kang WY, Roh B, Kim H (2023) Large language models are temporal and causal reasoners for video question answering. In: Proceedings of the 2023 Conference on Empirical Methods in Natural Language Processing (EMNLP), pp 4300--4316

\bibitem[{Kojima et~al.(2022)Kojima, Gu, Reid, Matsuo, and Iwasawa}]{kojima2022large}
Kojima T, Gu SS, Reid M, Matsuo Y, Iwasawa Y (2022) Large language models are zero-shot reasoners. Advances in neural information processing systems (NeurIPS) 35:22199--22213

\bibitem[{Le et~al.(2020)Le, Le, Venkatesh, and Tran}]{le2020hierarchical}
Le TM, Le V, Venkatesh S, Tran T (2020) Hierarchical conditional relation networks for video question answering. In: Proceedings of the IEEE/CVF Conference on Computer Vision and Pattern Recognition (CVPR), pp 9972--9981

\bibitem[{Lei et~al.(2021)Lei, Li, Zhou, Gan, Berg, Bansal, and Liu}]{lei2021less}
Lei J, Li L, Zhou L, Gan Z, Berg TL, Bansal M, Liu J (2021) Less is more: Clipbert for video-and-language learning via sparse sampling. In: Proceedings of the IEEE/CVF Conference on Computer Vision and Pattern Recognition (CVPR), pp 7331--7341

\bibitem[{Lei et~al.(2023)Lei, Berg, and Bansal}]{lei2023revealing}
Lei J, Berg T, Bansal M (2023) Revealing single frame bias for video-and-language learning. In: Proceedings of the 61st Annual Meeting of the Association for Computational Linguistics (Volume 1: Long Papers), pp 487--507

\bibitem[{Li et~al.(2023{\natexlab{a}})Li, Li, Savarese, and Hoi}]{li2023blip}
Li J, Li D, Savarese S, Hoi S (2023{\natexlab{a}}) Blip-2: Bootstrapping language-image pre-training with frozen image encoders and large language models. In: International conference on machine learning (ICML), PMLR, pp 19730--19742

\bibitem[{Li et~al.(2023{\natexlab{b}})Li, He, Wang, Li, Wang, Luo, Wang, Wang, and Qiao}]{li2023videochat}
Li K, He Y, Wang Y, Li Y, Wang W, Luo P, Wang Y, Wang L, Qiao Y (2023{\natexlab{b}}) Videochat: Chat-centric video understanding. arXiv preprint arXiv:230506355

\bibitem[{Li et~al.(2023{\natexlab{c}})Li, Wang, Li, Wang, He, Wang, and Qiao}]{li2023unmasked}
Li K, Wang Y, Li Y, Wang Y, He Y, Wang L, Qiao Y (2023{\natexlab{c}}) Unmasked teacher: Towards training-efficient video foundation models. In: Proceedings of the IEEE/CVF International Conference on Computer Vision (ICCV)

\bibitem[{Li et~al.(2024{\natexlab{a}})Li, Wang, He, Li, Wang, Liu, Wang, Xu, Chen, Luo et~al.}]{li2023mvbench}
Li K, Wang Y, He Y, Li Y, Wang Y, Liu Y, Wang Z, Xu J, Chen G, Luo P, et~al. (2024{\natexlab{a}}) Mvbench: A comprehensive multi-modal video understanding benchmark. In: Proceedings of the IEEE/CVF Conference on Computer Vision and Pattern Recognition (CVPR), pp 22195--22206

\bibitem[{Li et~al.(2020)Li, Chen, Cheng, Gan, Yu, and Liu}]{li2020hero}
Li L, Chen YC, Cheng Y, Gan Z, Yu L, Liu J (2020) Hero: Hierarchical encoder for video+ language omni-representation pre-training. In: Proceedings of the 2020 Conference on Empirical Methods in Natural Language Processing (EMNLP), pp 2046--2065

\bibitem[{Li et~al.(2023{\natexlab{d}})Li, Li, Ren, Liu, Liu, Gao, Sun, and Hou}]{li2023vitatecs}
Li S, Li L, Ren S, Liu Y, Liu Y, Gao R, Sun X, Hou L (2023{\natexlab{d}}) Vitatecs: A diagnostic dataset for temporal concept understanding of video-language models. arXiv preprint arXiv:231117404

\bibitem[{Li et~al.(2022)Li, Wang, Xiao, Ji, and Chua}]{li2022invariant}
Li Y, Wang X, Xiao J, Ji W, Chua TS (2022) Invariant grounding for video question answering. In: IEEE/CVF Conference on Computer Vision and Pattern Recognition (CVPR), pp 2928--2937

\bibitem[{Li et~al.(2023{\natexlab{e}})Li, Wang, Xiao, Ji, and Chua}]{li2023transformer}
Li Y, Wang X, Xiao J, Ji W, Chua TS (2023{\natexlab{e}}) Transformer-empowered invariant grounding for video question answering. IEEE Transactions on Pattern Analysis and Machine Intelligence

\bibitem[{Li et~al.(2023{\natexlab{f}})Li, Xiao, Feng, Wang, and Chua}]{li2023discovering}
Li Y, Xiao J, Feng C, Wang X, Chua TS (2023{\natexlab{f}}) Discovering spatio-temporal rationales for video question answering. In: IEEE/CVF International Conference on Computer Vision (ICCV), pp 13869--13878

\bibitem[{Li et~al.(2024{\natexlab{b}})Li, Chen, Hu, Wang, Shi, and Zhang}]{li2024videovista}
Li Y, Chen X, Hu B, Wang L, Shi H, Zhang M (2024{\natexlab{b}}) Videovista: A versatile benchmark for video understanding and reasoning. arXiv preprint arXiv:240611303

\bibitem[{Li et~al.(2024{\natexlab{c}})Li, Wang, and Jia}]{li2023llama}
Li Y, Wang C, Jia J (2024{\natexlab{c}}) Llama-vid: An image is worth 2 tokens in large language models. In: European Conference on Computer Vision (ECCV), Springer, pp 323--340

\bibitem[{Lin et~al.(2023)Lin, Zhu, Ye, Ning, Jin, and Yuan}]{lin2023video}
Lin B, Zhu B, Ye Y, Ning M, Jin P, Yuan L (2023) Video-llava: Learning united visual representation by alignment before projection. arXiv preprint arXiv:231110122

\bibitem[{Liu et~al.(2024{\natexlab{a}})Liu, Dong, Wang, Rao, Tang, Ma, and Krishna}]{liu2024coarse}
Liu B, Dong Y, Wang Y, Rao Y, Tang Y, Ma WC, Krishna R (2024{\natexlab{a}}) Coarse correspondence elicit 3d spacetime understanding in multimodal language model. arXiv preprint arXiv:240800754

\bibitem[{Liu et~al.(2023)Liu, Li, Wu, and Lee}]{liu2023visual}
Liu H, Li C, Wu Q, Lee YJ (2023) Visual instruction tuning. Advances in neural information processing systems (NeurIPS) 36

\bibitem[{Liu et~al.(2019)Liu, Ott, Goyal, Du, Joshi, Chen, Levy, Lewis, Zettlemoyer, and Stoyanov}]{liu2019roberta}
Liu Y, Ott M, Goyal N, Du J, Joshi M, Chen D, Levy O, Lewis M, Zettlemoyer L, Stoyanov V (2019) Roberta: A robustly optimized bert pretraining approach. arXiv preprint arXiv:190711692

\bibitem[{Liu et~al.(2024{\natexlab{b}})Liu, Li, Liu, Wang, Ren, Li, Chen, Sun, and Hou}]{liu2024tempcompass}
Liu Y, Li S, Liu Y, Wang Y, Ren S, Li L, Chen S, Sun X, Hou L (2024{\natexlab{b}}) Tempcompass: Do video llms really understand videos? arXiv preprint arXiv:240300476

\bibitem[{Liu et~al.(2022)Liu, Ning, Cao, Wei, Zhang, Lin, and Hu}]{liu2022video}
Liu Z, Ning J, Cao Y, Wei Y, Zhang Z, Lin S, Hu H (2022) Video swin transformer. In: Proceedings of the IEEE/CVF conference on computer vision and pattern recognition (CVPR), pp 3202--3211

\bibitem[{Maaz et~al.(2023)Maaz, Rasheed, Khan, and Khan}]{maaz2023video}
Maaz M, Rasheed H, Khan S, Khan FS (2023) Video-chatgpt: Towards detailed video understanding via large vision and language models. arXiv preprint arXiv:230605424

\bibitem[{Majumdar et~al.(2024)Majumdar, Ajay, Zhang, Putta, Yenamandra, Henaff, Silwal, Mcvay, Maksymets, Arnaud et~al.}]{majumdar2024openeqa}
Majumdar A, Ajay A, Zhang X, Putta P, Yenamandra S, Henaff M, Silwal S, Mcvay P, Maksymets O, Arnaud S, et~al. (2024) Openeqa: Embodied question answering in the era of foundation models. In: Proceedings of the IEEE/CVF Conference on Computer Vision and Pattern Recognition (CVPR), pp 16488--16498

\bibitem[{Mangalam et~al.(2023)Mangalam, Akshulakov, and Malik}]{mangalam2023egoschema}
Mangalam K, Akshulakov R, Malik J (2023) Egoschema: A diagnostic benchmark for very long-form video language understanding. In: The 37th Conference on Neural Information Processing Systems (NeurIPS) Track on Datasets and Benchmarks.

\bibitem[{Min et~al.(2024)Min, Buch, Nagrani, Cho, and Schmid}]{min2024morevqa}
Min J, Buch S, Nagrani A, Cho M, Schmid C (2024) Morevqa: Exploring modular reasoning models for video question answering. In: Proceedings of the IEEE/CVF Conference on Computer Vision and Pattern Recognition (CVPR), pp 13235--13245

\bibitem[{Niu et~al.(2021)Niu, Tang, Zhang, Lu, Hua, and Wen}]{niu2021counterfactual}
Niu Y, Tang K, Zhang H, Lu Z, Hua XS, Wen JR (2021) Counterfactual vqa: A cause-effect look at language bias. In: Proceedings of the IEEE/CVF Conference on Computer Vision and Pattern Recognition (CVPR), pp 12700--12710

\bibitem[{OpenAI(2023)}]{openai2023gpt4}
OpenAI (2023) Gpt-4 technical report. \eprint{2303.08774}

\bibitem[{P{\u{a}}tr{\u{a}}ucean et~al.(2023)P{\u{a}}tr{\u{a}}ucean, Smaira, Gupta, Continente, Markeeva, Banarse, Koppula, Heyward, Malinowski, Yang et~al.}]{puatruaucean2023perception}
P{\u{a}}tr{\u{a}}ucean V, Smaira L, Gupta A, Continente AR, Markeeva L, Banarse D, Koppula S, Heyward J, Malinowski M, Yang Y, et~al. (2023) Perception test: A diagnostic benchmark for multimodal video models. In: The 37th Conference on Neural Information Processing Systems (NeurIPS) Track on Datasets and Benchmarks.

\bibitem[{Radford et~al.(2021)Radford, Kim, Hallacy, Ramesh, Goh, Agarwal, Sastry, Askell, Mishkin, Clark et~al.}]{radford2021learning}
Radford A, Kim JW, Hallacy C, Ramesh A, Goh G, Agarwal S, Sastry G, Askell A, Mishkin P, Clark J, et~al. (2021) Learning transferable visual models from natural language supervision. In: International Conference on Machine Learning (ICML), PMLR, pp 8748--8763

\bibitem[{Seo et~al.(2021)Seo, Nagrani, and Schmid}]{seo2021look}
Seo PH, Nagrani A, Schmid C (2021) Look before you speak: Visually contextualized utterances. In: Proceedings of the IEEE/CVF Conference on Computer Vision and Pattern Recognition (CVPR), pp 16877--16887

\bibitem[{Shah et~al.(2019)Shah, Chen, Rohrbach, and Parikh}]{shah2019cycle}
Shah M, Chen X, Rohrbach M, Parikh D (2019) Cycle-consistency for robust visual question answering. In: Proceedings of the IEEE/CVF Conference on Computer Vision and Pattern Recognition (CVPR), pp 6649--6658

\bibitem[{Shang et~al.(2024)Shang, You, Subramanian, Darrell, and Herzig}]{shang2024traveler}
Shang C, You A, Subramanian S, Darrell T, Herzig R (2024) Traveler: A multi-lmm agent framework for video question-answering. arXiv preprint arXiv:240401476

\bibitem[{Sun et~al.(2019)Sun, Myers, Vondrick, Murphy, and Schmid}]{sun2019videobert}
Sun C, Myers A, Vondrick C, Murphy K, Schmid C (2019) Videobert: A joint model for video and language representation learning. In: Proceedings of the IEEE/CVF International Conference on Computer Vision (ICCV), pp 7464--7473

\bibitem[{Sun et~al.(2023)Sun, Fang, Wu, Wang, and Cao}]{sun2023eva}
Sun Q, Fang Y, Wu L, Wang X, Cao Y (2023) Eva-clip: Improved training techniques for clip at scale. arXiv preprint arXiv:230315389

\bibitem[{Sur{\'\i}s et~al.(2023)Sur{\'\i}s, Menon, and Vondrick}]{suris2023vipergpt}
Sur{\'\i}s D, Menon S, Vondrick C (2023) Vipergpt: Visual inference via python execution for reasoning. In: Proceedings of the IEEE/CVF International Conference on Computer Vision, pp 11888--11898

\bibitem[{Tang et~al.(2023)Tang, Bi, Xu, Song, Liang, Wang, Zhang, An, Lin, Zhu et~al.}]{tang2023video}
Tang Y, Bi J, Xu S, Song L, Liang S, Wang T, Zhang D, An J, Lin J, Zhu R, et~al. (2023) Video understanding with large language models: A survey. arXiv preprint arXiv:231217432

\bibitem[{Team et~al.(2023)Team, Anil, Borgeaud, Wu, Alayrac, Yu, Soricut, Schalkwyk, Dai, Hauth et~al.}]{gemini2023}
Team G, Anil R, Borgeaud S, Wu Y, Alayrac JB, Yu J, Soricut R, Schalkwyk J, Dai AM, Hauth A, et~al. (2023) Gemini: a family of highly capable multimodal models. arXiv preprint arXiv:231211805

\bibitem[{Touvron et~al.(2023)Touvron, Lavril, Izacard, Martinet, Lachaux, Lacroix, Rozi{\`e}re, Goyal, Hambro, Azhar et~al.}]{touvron2023llama}
Touvron H, Lavril T, Izacard G, Martinet X, Lachaux MA, Lacroix T, Rozi{\`e}re B, Goyal N, Hambro E, Azhar F, et~al. (2023) Llama: Open and efficient foundation language models. arXiv preprint arXiv:230213971

\bibitem[{Wang et~al.(2024{\natexlab{a}})Wang, Zhang, Zohar, and Yeung-Levy}]{wang2024videoagent}
Wang X, Zhang Y, Zohar O, Yeung-Levy S (2024{\natexlab{a}}) Videoagent: Long-form video understanding with large language model as agent. European Conference on Computer Vision (ECCV)

\bibitem[{Wang et~al.(2024{\natexlab{b}})Wang, Yu, Stengel-Eskin, Yoon, Cheng, Bertasius, and Bansal}]{wang2024videotree}
Wang Z, Yu S, Stengel-Eskin E, Yoon J, Cheng F, Bertasius G, Bansal M (2024{\natexlab{b}}) Videotree: Adaptive tree-based video representation for llm reasoning on long videos. arXiv preprint arXiv:240519209

\bibitem[{Wei et~al.(2022)Wei, Wang, Schuurmans, Bosma, Xia, Chi, Le, Zhou et~al.}]{wei2022chain}
Wei J, Wang X, Schuurmans D, Bosma M, Xia F, Chi E, Le QV, Zhou D, et~al. (2022) Chain-of-thought prompting elicits reasoning in large language models. Advances in neural information processing systems (NeurIPS) 35:24824--24837

\bibitem[{Xiao et~al.(2021)Xiao, Shang, Yao, and Chua}]{xiao2021next}
Xiao J, Shang X, Yao A, Chua TS (2021) Next-qa: Next phase of question-answering to explaining temporal actions. In: Proceedings of the IEEE/CVF Conference on Computer Vision and Pattern Recognition (CVPR), pp 9777--9786

\bibitem[{Xiao et~al.(2022{\natexlab{a}})Xiao, Yao, Liu, Li, Ji, and Chua}]{xiao2022hqga}
Xiao J, Yao A, Liu Z, Li Y, Ji W, Chua TS (2022{\natexlab{a}}) Video as conditional graph hierarchy for multi-granular question answering. In: Proceedings of the AAAI Conference on Artificial Intelligence (AAAI), vol~36, pp 2804--2812

\bibitem[{Xiao et~al.(2022{\natexlab{b}})Xiao, Zhou, Chua, and Yan}]{xiao2022vgt}
Xiao J, Zhou P, Chua TS, Yan S (2022{\natexlab{b}}) Video graph transformer for video question answering. In: European Conference on Computer Vision (ECCV), Springer, pp 39--58

\bibitem[{Xiao et~al.(2023)Xiao, Zhou, Yao, Li, Hong, Yan, and Chua}]{xiao2023covgt}
Xiao J, Zhou P, Yao A, Li Y, Hong R, Yan S, Chua TS (2023) Contrastive video question answering via video graph transformer. IEEE Transactions on Pattern Analysis and Machine Intelligence (T-PAMI) 45(11):13265--13280, \doi{10.1109/TPAMI.2023.3292266}

\bibitem[{Xiao et~al.(2024)Xiao, Yao, Li, and Chua}]{xiao2023can}
Xiao J, Yao A, Li Y, Chua TS (2024) Can i trust your answer? visually grounded video question answering. In: Proceedings of the IEEE/CVF Conference on Computer Vision and Pattern Recognition (CVPR), pp 13204--13214

\bibitem[{Xu et~al.(2017)Xu, Zhao, Xiao, Wu, Zhang, He, and Zhuang}]{xu2017video}
Xu D, Zhao Z, Xiao J, Wu F, Zhang H, He X, Zhuang Y (2017) Video question answering via gradually refined attention over appearance and motion. In: Proceedings of the 25th ACM international conference on Multimedia, pp 1645--1653

\bibitem[{Yang et~al.(2021)Yang, Miech, Sivic, Laptev, and Schmid}]{yang2021just}
Yang A, Miech A, Sivic J, Laptev I, Schmid C (2021) Just ask: Learning to answer questions from millions of narrated videos. In: Proceedings of the IEEE/CVF International Conference on Computer Vision (ICCV), pp 1686--1697

\bibitem[{Yang et~al.(2022)Yang, Miech, Sivic, Laptev, and Schmid}]{yang2022zero}
Yang A, Miech A, Sivic J, Laptev I, Schmid C (2022) Zero-shot video question answering via frozen bidirectional language models. Advances in Neural Information Processing Systems (NeurIPS) 35:124--141

\bibitem[{Yu et~al.(2023)Yu, Cho, Yadav, and Bansal}]{yu2023self}
Yu S, Cho J, Yadav P, Bansal M (2023) Self-chained image-language model for video localization and question answering. In: The 37th Conference on Neural Information Processing Systems (NeurIPS)

\bibitem[{Yu et~al.(2019)Yu, Xu, Yu, Yu, Zhao, Zhuang, and Tao}]{yu2019activitynet}
Yu Z, Xu D, Yu J, Yu T, Zhao Z, Zhuang Y, Tao D (2019) Activitynet-qa: A dataset for understanding complex web videos via question answering. In: Proceedings of the AAAI Conference on Artificial Intelligence (AAAI), vol~33, pp 9127--9134

\bibitem[{Zeng et~al.(2023)Zeng, Attarian, Choromanski, Wong, Welker, Tombari, Purohit, Ryoo, Sindhwani, Lee, and Florence}]{zengsocratic}
Zeng A, Attarian M, Choromanski KM, Wong A, Welker S, Tombari F, Purohit A, Ryoo MS, Sindhwani V, Lee VV Johnny, Florence P (2023) Socratic models: Composing zero-shot multimodal reasoning with language. In: The Eleventh International Conference on Learning Representations (ICLR)

\bibitem[{Zhang et~al.(2023{\natexlab{a}})Zhang, Lu, Islam, Wang, Yu, Bansal, and Bertasius}]{zhang2023simple}
Zhang C, Lu T, Islam MM, Wang Z, Yu S, Bansal M, Bertasius G (2023{\natexlab{a}}) A simple llm framework for long-range video question-answering. arXiv preprint arXiv:231217235

\bibitem[{Zhang et~al.(2023{\natexlab{b}})Zhang, Li, and Bing}]{zhang2023videollama}
Zhang H, Li X, Bing L (2023{\natexlab{b}}) Video-llama: An instruction-tuned audio-visual language model for video understanding. In: Proceedings of the 2023 Conference on Empirical Methods in Natural Language Processing: System Demonstrations, pp 543--553

\bibitem[{Zhang et~al.(2024{\natexlab{a}})Zhang, Zhang, Li, Zeng, Yang, Zhang, Wang, Tan, Li, and Liu}]{zhang2024long}
Zhang P, Zhang K, Li B, Zeng G, Yang J, Zhang Y, Wang Z, Tan H, Li C, Liu Z (2024{\natexlab{a}}) Long context transfer from language to vision. arXiv preprint arXiv:240616852

\bibitem[{Zhang et~al.(2023{\natexlab{c}})Zhang, Han, Liu, Gao, Zhou, Hu, Yan, Lu, Li, and Qiao}]{zhang2023llamaadapter}
Zhang R, Han J, Liu C, Gao P, Zhou A, Hu X, Yan S, Lu P, Li H, Qiao Y (2023{\natexlab{c}}) Llama-adapter: Efficient fine-tuning of language models with zero-init attention. arXiv preprint arXiv:230316199

\bibitem[{Zhang et~al.(2024{\natexlab{b}})Zhang, Han, Zhou, Hu, Yan, Lu, Li, Gao, and Qiao}]{zhang2023llama}
Zhang R, Han J, Zhou A, Hu X, Yan S, Lu P, Li H, Gao P, Qiao Y (2024{\natexlab{b}}) Llama-adapter: Efficient fine-tuning of language models with zero-init attention. ICLR

\bibitem[{Zhang et~al.(2023{\natexlab{d}})Zhang, Zhang, and Xu}]{zhang2023reducing}
Zhang X, Zhang F, Xu C (2023{\natexlab{d}}) Reducing vision-answer biases for multiple-choice vqa. IEEE Transactions on Image Processing (TIP) pp 4621--4634

\bibitem[{Zhang et~al.(2024{\natexlab{c}})Zhang, Zhang, and Xu}]{zhang2023next}
Zhang X, Zhang F, Xu C (2024{\natexlab{c}}) Next-ood: Overcoming dual multiple-choice vqa biases. IEEE Transactions on Pattern Analysis and Machine Intelligence (T-PAMI) 46(4):1913--1931, \doi{10.1109/TPAMI.2023.3269429}

\bibitem[{Zhang et~al.(2024{\natexlab{d}})Zhang, Huang, Ma, Li, Luo, Xie, Qin, Luo, Li, Liu et~al.}]{zhang2023recognize}
Zhang Y, Huang X, Ma J, Li Z, Luo Z, Xie Y, Qin Y, Luo T, Li Y, Liu S, et~al. (2024{\natexlab{d}}) Recognize anything: A strong image tagging model. In: Proceedings of the IEEE/CVF Conference on Computer Vision and Pattern Recognition (CVPR), pp 1724--1732

\bibitem[{Zhang et~al.(2024{\natexlab{e}})Zhang, Li, Liu, Lee, Gui, Fu, Feng, Liu, and Li}]{zhang2024llavanextvideo}
Zhang Y, Li B, Liu h, Lee Yj, Gui L, Fu D, Feng J, Liu Z, Li C (2024{\natexlab{e}}) Llava-next: A strong zero-shot video understanding model. \urlprefix\url{https://llava-vl.github.io/blog/2024-04-30-llava-next-video/}

\bibitem[{Zhang et~al.(2023{\natexlab{e}})Zhang, Zhang, Li, Zhao, Karypis, and Smola}]{zhang2023multimodal}
Zhang Z, Zhang A, Li M, Zhao H, Karypis G, Smola A (2023{\natexlab{e}}) Multimodal chain-of-thought reasoning in language models. arXiv preprint arXiv:230200923

\bibitem[{Zhao et~al.(2023)Zhao, Misra, Kr{\"a}henb{\"u}hl, and Girdhar}]{zhao2023learning}
Zhao Y, Misra I, Kr{\"a}henb{\"u}hl P, Girdhar R (2023) Learning video representations from large language models. In: Proceedings of the IEEE/CVF Conference on Computer Vision and Pattern Recognition (CVPR), pp 6586--6597

\bibitem[{Zhong et~al.(2022)Zhong, Xiao, Ji, Li, Deng, and Chua}]{zhong2022video}
Zhong Y, Xiao J, Ji W, Li Y, Deng W, Chua TS (2022) Video question answering: Datasets, algorithms and challenges. In: Proceedings of the 2022 Conference on Empirical Methods in Natural Language Processing (EMNLP), pp 6439--6455

\bibitem[{Zhu et~al.(2023)Zhu, Lin, Ning, Yan, Cui, Wang, Pang, Jiang, Zhang, Li et~al.}]{zhu2023languagebind}
Zhu B, Lin B, Ning M, Yan Y, Cui J, Wang H, Pang Y, Jiang W, Zhang J, Li Z, et~al. (2023) Languagebind: Extending video-language pretraining to n-modality by language-based semantic alignment. arXiv preprint arXiv:231001852

\end{thebibliography}
